\documentclass{article}

\usepackage[svgnames]{xcolor}

\usepackage{iclr2023_conference,times}

\usepackage[T1]{fontenc}
\usepackage[utf8]{inputenc}
\usepackage[english]{babel}
\usepackage[ruled,noend]{algorithm2e}
\usepackage{bm}
\usepackage{dsfont}
\usepackage{pifont}
\usepackage{etoolbox}
\usepackage[pagebackref=true]{hyperref}
\usepackage{url}
\usepackage{graphicx}
\usepackage{xcolor}
\usepackage{mathtools}
\usepackage{amssymb}
\usepackage{commath}
\usepackage{dsfont}
\usepackage{mathrsfs}
\usepackage[short]{optidef}
\usepackage{stmaryrd}
\usepackage{nicefrac}
\usepackage{bigdelim}
\usepackage{subfig}
\usepackage{xfrac}
\usepackage[separate-uncertainty=true]{siunitx}
\usepackage{microtype}
\usepackage{soul}
\usepackage{cases}
\usepackage{import}
\usepackage{booktabs}
\usepackage{transparent}
\usepackage{multirow}
\usepackage{rotating}
\usepackage{multido}
\usepackage{makecell}
\usepackage{graphics}
\usepackage{natbib}
\usepackage{wrapfig}
\usepackage{xspace}
\usepackage{comment}
\usepackage[capitalize,noabbrev]{cleveref}
\usepackage[inline,shortlabels]{enumitem}
\usepackage[normalem]{ulem}
\usepackage{colortbl}
\usepackage{float}

\renewcommand*{\backrefalt}[4]{%
    \ifcase #1 {(Not cited.)}%
    \or        {(cit.\ on p.~#2)}%
    \else      {(cit.\ on pp.~#2)}%
    \fi
}

\definecolor{cpink}{HTML}{FCCDE5}
\definecolor{cred}{HTML}{FFC2BA}
\definecolor{cyellow}{HTML}{FFFFB3}
\definecolor{cblue}{HTML}{B9DEFF}
\definecolor{cgreen}{HTML}{D7F3E7}
\definecolor{cneutral}{HTML}{CFCFCF}

\DeclareRobustCommand{\cyellow}[1]{\sethlcolor{cyellow}{\textbf{\hl{~#1~}}}}
\DeclareRobustCommand{\cgreen}[1]{\sethlcolor{cgreen}{\textbf{\hl{~#1~}}}}
\DeclareRobustCommand{\cred}[1]{\sethlcolor{cred}{\textbf{\hl{~#1~}}}}

\newcommand*\X{\mathcal{X}}

\let\originalleft\left
\let\originalright\right
\renewcommand{\left}{\mathopen{}\mathclose\bgroup\originalleft}
\renewcommand{\right}{\aftergroup\egroup\originalright}

\def\eqref#1{eq.~(\ref{#1})}
\def\Eqref#1{Eq.~(\ref{#1})}

\def\1{\bm{1}}
\newcommand{\ours}{\textsc{DINo}\xspace}

\DeclareMathAlphabet{\mathsfit}{\encodingdefault}{\sfdefault}{m}{sl}
\SetMathAlphabet{\mathsfit}{bold}{\encodingdefault}{\sfdefault}{bx}{n}

\def\gD{{\mathcal{D}}}

\def\gT{{\mathcal{T}}}
\def\gU{{\mathcal{U}}}
\def\gV{{\mathcal{V}}}

\def\gX{{\mathcal{X}}}

\newcommand{\R}{\mathbb{R}}

\newcommand{\tr}{\text{tr}}
\newcommand{\ts}{\text{ts}}

\DeclareMathOperator*{\argmin}{arg\,min}

\newcommand*{\eg}{e.g.\@\xspace}
\newcommand*{\sut}{s.t.\@\xspace}
\newcommand*{\ie}{i.e.\@\xspace}

\newcommand*{\wrt}{w.r.t.\@\xspace}
\newcommand*{\aka}{a.k.a.\@\xspace}

\newcommand*{\cf}{cf.\@\xspace}

\newcommand*{\resp}{resp.\@\xspace}

\let\up\textsuperscript

\DeclarePairedDelimiter\parentheses{(}{)}

\DeclarePairedDelimiter\brackets{[}{]}

\DeclarePairedDelimiter\lrbrackets{\llbracket}{\rrbracket}
\DeclarePairedDelimiter\euclideannorm{\|}{\|}

\DeclarePairedDelimiter\rightvert{.}{\vert}

\DeclarePairedDelimiterX{\midx}[2]{(}{)}{#1\;\delimsize\vert\;#2}
\DeclarePairedDelimiterX{\midbracesx}[2]{\{}{\}}{#1\;\delimsize\vert\;#2}
\DeclarePairedDelimiterX{\parallelx}[2]{(}{)}{#1\;\delimsize\|\;#2}

\newcommand{\app}[2]{#1 \parentheses*{#2}}

\newcommand{\cmark}{\color{ForestGreen} \ding{51}}
\newcommand{\xmark}{\color{Red} \ding{55}}

\SetKwComment{Comment}{/* }{ */}

\robustify\uline
\def\Uline#1{#1\llap{\uline{\phantom{\num{#1}}}}}

\newrobustcmd{\B}{\fontseries{b}\selectfont}

\definecolor{mydarkblue}{rgb}{0,0.08,0.45}
\hypersetup{
    colorlinks=true,
    linkcolor=mydarkblue,
    citecolor=mydarkblue,
    filecolor=mydarkblue,
    urlcolor=mydarkblue,
    pdfview=FitH
}

\everypar{\looseness=-1}
\usepackage{titlesec}
\titlespacing{\paragraph}{%
  0pt}{
  0pt}{
  1em}

\title{Continuous PDE Dynamics Forecasting with Implicit Neural Representations}

\author{%
  Yuan Yin\thanks{Equal contribution}\up{~~1}\hfill Matthieu Kirchmeyer${}^*$\up{\!1,2}\hfill Jean-Yves Franceschi${}^*$\up{\!2}\\
  \textbf{Alain Rakotomamonjy\up{2}\hfill Patrick Gallinari\up{1,2}} \\
  \up{1}Sorbonne Université, CNRS, ISIR, F-75005 Paris, France ~~~~~ \up{2}Criteo AI Lab, Paris
}

\newcommand{\rev}[1]{#1}

\iclrfinalcopy

\robustify\cellcolor

\begin{document}

    \maketitle

    \begin{abstract}
        Effective data-driven PDE forecasting methods often rely on fixed spatial and\,/\,or temporal discretizations.
        This raises limitations in real-world applications like weather prediction where flexible extrapolation at arbitrary spatiotemporal locations is required.
        We address this problem by introducing a new data-driven approach, \ours, that models a PDE's flow with continuous-time dynamics of spatially continuous functions.
        This is achieved by embedding spatial observations independently of their discretization via Implicit Neural Representations in a small latent space temporally driven by a learned ODE.
        This separate and flexible treatment of time and space makes \ours the first data-driven model to combine the following advantages.
        It extrapolates at arbitrary spatial and temporal locations; it can learn from sparse irregular grids or manifolds; at test time, it generalizes to new grids or resolutions.
        \ours outperforms alternative neural PDE forecasters in a variety of challenging generalization scenarios on representative PDE systems.
    \end{abstract}

    \section{Introduction}
\label{sec:intro}
Modeling the dynamics and predicting the temporal evolution of physical phenomena is paramount in many fields, \eg climate modeling, biology, fluid mechanics and energy \citep{Willard2022}.
Classical solutions rely on a well-established physical paradigm: the evolution is described by differential equations derived from physical first principles, and then solved using numerical analysis tools, \eg finite elements, finite volumes or spectral methods \citep{Olver2014}.
The availability of large amounts of data from observations or simulations has motivated data-driven approaches to this problem \citep{Brunton2022}, leading to a rapid development of the field with deep learning methods.
The main motivations for this research track include developing surrogate or reduced order models that can approximate high-fidelity full order models at reduced computational costs \citep{Kochkov2021}, complementing classical solvers, \eg to account for additional components of the dynamics \citep{Yin2021}, or improving low fidelity models \citep{Avila2020}.

Most of these attempts rely on workhorses of deep learning like CNNs \citep{Ayed2020} or GNNs \citep{Li2020,Pfaff2021,Brandstetter2022}.
They all require prior space discretization either on regular or irregular grids, such that they only capture the dynamics on the train grid and cannot generalize outside it.
Neural operators, a recent trend, learn mappings between function spaces \citep{Li2021, Lu2021} and thus alleviate some limitations of prior discretization approaches.
Yet, they still rely on fixed grid discretization for training and inference: \eg, regular grids for \cite{Li2021} or a free-form but predetermined grid for \cite{Lu2021}.
Hence, the number and\,/\,or location of the sensors has to be fixed across train and test which is restrictive in many situations \citep{Prasthofer2022}.
Mesh-agnostic approaches for solving canonical PDEs (Partial Differential Equations) are another trend \citep{Raissi2019, Sirignano2018}.
In contrast to physics-agnostic grid-based approaches, they aim at solving a known PDE as usual solvers do, and cannot cope with unknown dynamics.
This idea was concurrently developed for computer graphics, \eg for learning 3D shapes \citep{Sitzmann2019,Mildenhall2020, Tancik2020} and coined as Implicit Neural Representations (INRs).
When used as solvers, these methods can only tackle a single initial value problem and are not designed for long-term forecasting outside the training horizon.

Because of these limitations, none of the above approaches can handle situations encountered in many practical applications such as:
    different geometries, \eg phenomena lying on a Euclidean plane or an Earth-like sphere;
    variable sampling, \eg irregular observation grids that may evolve at train and test time as in adaptive meshing \citep{Berger1984};
    scarce training data, \eg when observations are only available at a few spatiotemporal locations; multi-scale phenomena, \eg in large scale-dynamics systems as climate modeling, where integrating intertwined subgrid scales \aka the closure problem is ubiquitous \citep{Zanna2021}.
These considerations motivate the development of new machine learning models that improve existing approaches on several of these aspects.

In our work, we aim at forecasting PDE-based spatiotemporal physical processes with a versatile model tackling the aforementioned limitations.
We adopt an agnostic approach, \ie not assuming any prior knowledge on the physics.
We introduce \ours (Dynamics-aware Implicit Neural representations), a model operating continuously in space and time, with the following contributions.
\begin{description}[topsep=0pt, partopsep=0pt,leftmargin=0em, nosep]
    \item[Continuous flow learning.]
    \ours aims at learning the PDE's flow to forecast its solutions, in a continuous manner so that it can be trained on any spatial and temporal discretization and applied to another.
    To this end, \ours embeds spatial observations into a small latent space via INRs; then it models continuous-time evolution by a learned latent Ordinary Differential Equation (ODE).
    \item[Space-time separation.]
    To efficiently encode different sequences, we propose a novel INR parameterization, amplitude modulation, implementing a space-time separation of variables.
    This simplifies the learned dynamics, reduces the number of parameters and greatly improves performance.
    \item[Spatiotemporal versatility.]
    \ours combines the benefits of prior models; \cf \cref{tab:comparison}.
    It tackles new sequences via its amplitude modulation.
    Sequential modeling with an ODE makes it extrapolate to unseen times within or beyond the train horizon.
    Thanks to INRs' spatial flexibility, it generalizes to new grids or resolutions, predicts at arbitrary positions and handles sparse irregular grids or manifolds.
    \item[Empirical validation.] We demonstrate \ours's versatility and state-of-the-art performance versus prior neural PDE forecasters, representative of grid, operator and INR-based methods, via thorough experiments on challenging multi-dimensional PDEs in various spatiotemporal generalization settings.
\end{description}

\begin{table}
    \caption{
        \label{tab:comparison}
        Comparison of data-driven approaches to spatiotemporal PDE forecasting.
    }
    \centering
    \tiny
    \begin{tabular}{@{}r@{}l@{~~~~}l@{\quad}c@{\quad}c@{\quad}c@{\quad}c@{\quad}c@{\quad}c}
        \toprule
        \multicolumn{2}{c}{Model} & Reference & \makecell{1.\ PDE-agnostic \\ prediction on new \\ initial conditions} & \makecell{2. Train\,/ \\ test space grid \\ independence} & \makecell{3.\ Evaluation at \\ unobserved spa- \\ tial locations} & \makecell{4.\ Free-form spatial \\ domain (manifold, \\ irregular mesh)} & \makecell{5. Time\\ continuous } & \makecell{6. Time \\ extrapolation} \\
        \midrule
        \ldelim\{{2}{*}[Discrete] & NODE & \cite{Chen2018}
        & \cmark & \xmark & \xmark  & \xmark & \cmark & \cmark \\
        & MP-PDE & \cite{Brandstetter2022} & \cmark & \xmark & \xmark & \cmark & \xmark & \cmark \\
        \midrule
        \ldelim\{{2}{*}[Operator] & MNO & \cite{Li2021b} & \cmark & \cmark & \xmark  & \xmark & \xmark  & \cmark \\
        & DeepONet & \cite{Lu2021} & \cmark & \xmark & \cmark  & \cmark & \cmark & \xmark \\
        \midrule
        \ldelim\{{2}{*}[INRs] & PINNs & \cite{Raissi2019} & \xmark & \cmark & \cmark  & \cmark & \cmark & \xmark \\
        & \ours & Ours & \cmark & \cmark & \cmark & \cmark & \cmark & \cmark \\
        \bottomrule
    \end{tabular}
    \vspace{-1em}
\end{table}

    \section{Problem description}
\label{sec:tasks}

\begin{figure}
    \centering
    \includegraphics[width=\textwidth]{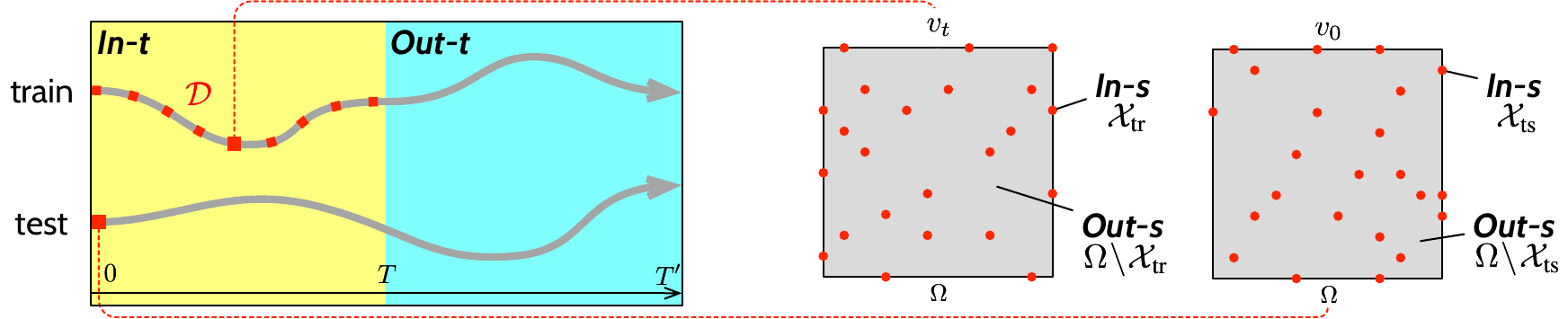}
    \caption[foo bar]{\label{fig:eval_notation}
        (Left)
        We represent time contexts.
        The \emph{train} trajectory consists of training snapshots ({\color{red} \mbox{\larger[-3]$\blacksquare$}}), observed in a train interval $[0,T]$ denoted \emph{In-t}.
        The line (\textcolor{gray}{\textbf{---}}) in continuation is a forecasting of this trajectory beyond \emph{In-t}, in $(T,T']$ denoted \emph{Out-t}.
        The line below (\textcolor{gray}{\textbf{---}}, \emph{test}) is a forecasting from a new initial condition $v_0$ ({\color{red} \mbox{\larger[-3]$\blacksquare$}}) on \emph{In-t} and \textit{Out-t}.
        (Middle and right) We illustrate spatial contexts.
        (Middle)
        Dots ({\color{red} $\bullet$}) correspond to the train observation grid $\gX_{\tr}$, denoted \emph{In-s}.
        \emph{Out-s} denotes the complementary domain $\Omega\setminus\gX_{\tr}$.
        (Right)
        New test observation grid $\gX_{\ts}$, used as an initial point for forecasting (left).
    }
    \vspace{-1.1em}
\end{figure}

\paragraph{Problem setting.}
We aim at modeling, via a data-driven approach, the temporal evolution of a continuous fully-observed \rev{deterministic} spatiotemporal phenomenon.
It is described by trajectories $v\colon\R\to\gV$ in a set $\varGamma$; we use $v_t \triangleq v(t) \in \gV$.
\rev{We focus on Initial Value Problems, where only $v_t$ at any time $t$ is required to infer $v_{t'}$ for $t'>t$.}
\rev{Hence,} trajectories share the same dynamics but differ by their initial condition $v_0\in\gV$.
$\R$ is the temporal domain and $\gV$ is the functional space of the form $\Omega\to\R^{n}$, where $\Omega \subset \R^p$ is a compact spatial domain and $n$ the number of observed values.
In other words, $v_t$ is a spatial function of $x \in \Omega$, with vectorial output $\app{v_t}{x} \in \R^n$; \cf examples of \cref{sec:exp_setting}.
To this end, we consider the setting illustrated in \Cref{fig:eval_notation}.
We observe a finite training set of trajectories $\gD$ with a free-form spatial observation grid $\gX_{\tr} \subset \Omega$ and on discrete times $t \in \gT \subset\brackets*{0, T}$.
At test time, we are only given a new initial condition $v_0$, with observed values $\rightvert*{v_0}_{\gX_{\ts}}$ on a new observation grid $\gX_{\ts}$, potentially different from $\gX_{\tr}$.
Inference is performed on both train and test trajectories given only the initial condition, on a new free-form grid $\gX' \subset \Omega$ and times $t \in \gT' \subset[0, T']$.
Inference grid $\gX'$ comprises observed positions (respectively $\gX_{\tr}$ and $\gX_{\ts}$ for train and test trajectories) and unobserved positions corresponding to \rev{spatial interpolation}.
Note that the inference temporal horizon is larger than the train one: $T<T'$.
For simplicity, \textit{In-s} refers to data in $\gX'$ on the observation grid ($\gX_{\tr}$ for \textit{train}\,/\,$\gX_{\ts}$ for \textit{test}), \textit{Out-s} to data in $\gX'$ outside the observation grid; \textit{In-t} refers to times within the train horizon $\gT\subset[0, T]$, and \textit{Out-t} to times in $\gT'\setminus\gT\subset(T, T']$, beyond $T$, up to inference horizon $T'$.

\paragraph{Evaluation scenarios.}
The desired properties in \Cref{sec:intro} call for spatiotemporally continuous forecasting models.
We select six criteria that our approach should meet; \cf column titles of \cref{tab:comparison}.
First, the model should be robust to the change of initial condition $v_0$, \ie generalize to \textit{test} trajectories (col.~1).
Second, it should extrapolate beyond the train conditions: in space, on a test observation grid that differs from the train one, \ie $\gX'=\gX_{\ts}\neq\gX_{\tr}$ (\textit{In-s}) (col.~2), and outside the observed train and test grid, \ie on $\gX'\setminus\gX_{\ts},\gX'\setminus\gX_{\tr}$ (\textit{Out-s}, col.~3); in time, between train snapshots (col.~5) and beyond the observed train horizon $T$ (\textit{Out-t}, col.~6).
Finally, it should adapt to free-form spatial domains, \ie to various geometries (\eg manifolds) or irregular grids (col.~4). See also \Cref{fig:eval_notation}.

\paragraph{Objective.}
To satisfy these requirements, we learn the flow $\Phi$ of the physical system:
\begin{align} \label{eq:flow}
    \textstyle \smash{\Phi\colon (\gV \times \R) \to \gV,} && \textstyle \smash{(v_{t}, \tau) \mapsto \Phi_{\tau}(v_{t}) = v_{t+\tau} \quad \forall v \in \varGamma, t\in\mathbb{R}.}
\end{align}
Learning the flow is a common strategy in sequential models to better generalize beyond the train time horizon.
Yet, so far, it has always been learned with discretized models, which poses generalization issues violating our requirements.
We describe these issues in \Cref{sec:related_work}.

    \section{Related work}
\label{sec:related_work}
We review current data-driven approaches for PDE modeling and the representative methods listed in \cref{tab:comparison}.
We express the forecasting rule using the notations in \Eqref{eq:flow}: $t$ is an arbitrary time; $\tau$ is an arbitrary time interval; $\delta t$ is a fixed, predetermined time interval (as a model hyperparameter).

\paragraph{Sequential discretized models.}
Most sequential dynamics models are learned on a fixed observed grid $\gX_{\tr}$ and use discretized models, \eg CNN or GNN to process the observations.
CNNs require observations on a regular grid but can be extended to irregular grids through interpolation \citep{Chae2021}.
GNNs are more flexible as they handle irregular grids, at an additional memory and computational cost.
Yet, prediction on new grids $\gX' \neq \gX_{\tr}$ fails experimentally for both CNNs and GNNs, as these discretized models are biased towards the training grid $\gX_{\tr}$, as we later show in \Cref{sec:experiments}.
We distinguish two types of temporal models which both extrapolate beyond the train horizon due to their sequential nature.
\begin{itemize*}
    \item Autoregressive models \bm{${v_{t}\vert}_{\gX} \mapsto {v_{t+\delta t}\vert}_{\gX}$} \citep{Long2018,deBezenac2018,Pfaff2021,Brandstetter2022} predict the sequence from $t$ only at fixed time increments $\delta t$ and not in between.
    \item Time-continuous extensions using numerical solvers \bm{$({v_{t}\vert}_{\gX}, \tau) \mapsto {v_{t+\tau}\vert}_{\gX}$} \citep{Yin2021,Iakovlev2021} solve this limitation as they provide a prediction at arbitrary times, thus remove dependency on the time discretization.
\end{itemize*}

\paragraph{Operator learning.}
Recently, operator-based models aim at finding a parameterized mapping between functions.
They define in theory space-continuous models.
First, neural operators \citep{Kovachki2021} attempt to replace standard convolution with continuous alternatives.
Fourier Neural Operator (FNO, \citealp{Li2021}) applies convolution in the spectral domain via Fast Fourier Transformation (FFT).
Graph Neural Operator (GNO, \citealp{Li2020}) performs convolution on a local interaction grid described by a graph.
Second, DeepONet \citep{Lu2021} uses a coordinate-based neural network to output a prediction at arbitrary time and space locations given a function observed on a fixed grid.
Three types of temporal models were used for operators with some limitations.
\begin{itemize*}[nosep,topsep=0pt]
    \item The standard approach, \bm{${v_0} \mapsto {v_t}$}, models the output at a given time $t\in[0,T]$ within the train horizon \citep{Li2020}.
    \item A sequential extension, \bm{${v_{t}} \mapsto v_{t+\delta t}$}, was proposed in \cite{Li2021b}.
    \item Finally, a time-continuous version \bm{${v_0} \mapsto (t\in[0,T] \mapsto v_{t})$} in DeepONet propose a solution at arbitrary time and space locations.
\end{itemize*}
The first and third approaches are not designed to generalize beyond the train horizon, \ie when $t>T$ as they are not sequential.
The second solves this limitation but is only able to predict solutions from $t$ at fixed time increments of $\delta t$ and not in-between.
Furthermore, all existing approaches make restrictive assumptions on the space discretization.
They lack flexibility when encoding spatial observations: FNO is limited to uniform Cartesian observation grids due to FFT, and while one concurrent follow-up alleviates this issue \citep{Li2022}, it still cannot perform predictions on unobserved spatial locations; GNO does not adapt well to changing observation grids as for the GNN-based models in the previous paragraph; DeepONet is limited to input observations on fixed observation locations.
The latter are chosen at random spatial positions but should remain fixed throughout training and testing.

\paragraph{Spatiotemporal INRs.}
Another class of models relies on coordinate-based neural networks, called Implicit Neural Representations \citep[INRs,][]{Sitzmann2019,Fathony2021,Tancik2020}.
These space-continuous models share a similar objective as operators, despite constituting a separate research field.
INRs for spatiotemporal data take time as an input along spatial coordinates.
Physics-informed neural networks (PINNs, \citealp{Raissi2019}) use this formulation to solve PDEs, yet are limited to a single known differential equation and a set of initial and boundary conditions.
\citet{Fresca2020} and \citet{Chen2023} combine INRs with reduced order models.
Extensions for multi-sequence learning, \eg for video generation \citep{Yu2022,Skorokhodov2022} or compression \citep{Chen2021}, learn a latent conditioning variable from an initial condition $v_0$, \ie take the form \bm{${v_0} \mapsto (t\in[0,T] \mapsto {v_t})$}.
Interestingly, these models can predict at an arbitrary time $t$ in the train horizon without unrolling a sequential model up to $t$.
Yet, as they only learn mappings from an initial condition ${v_0}$ to a function of time ${v_t}$ in the train domain, they fail to predict beyond train conditions, as we show in \cref{sec:experiments}.
\ours is a new instance of spatiotemporal INR which solves this limitation via a time-continuous dynamics model of the underlying flow, \bm{$({v_{t}}, \tau) \mapsto {v_{t+\tau}}$}.

    \section{Model}
\label{sec:model}
\begin{figure}
    \centering
    \includegraphics[width=\textwidth]{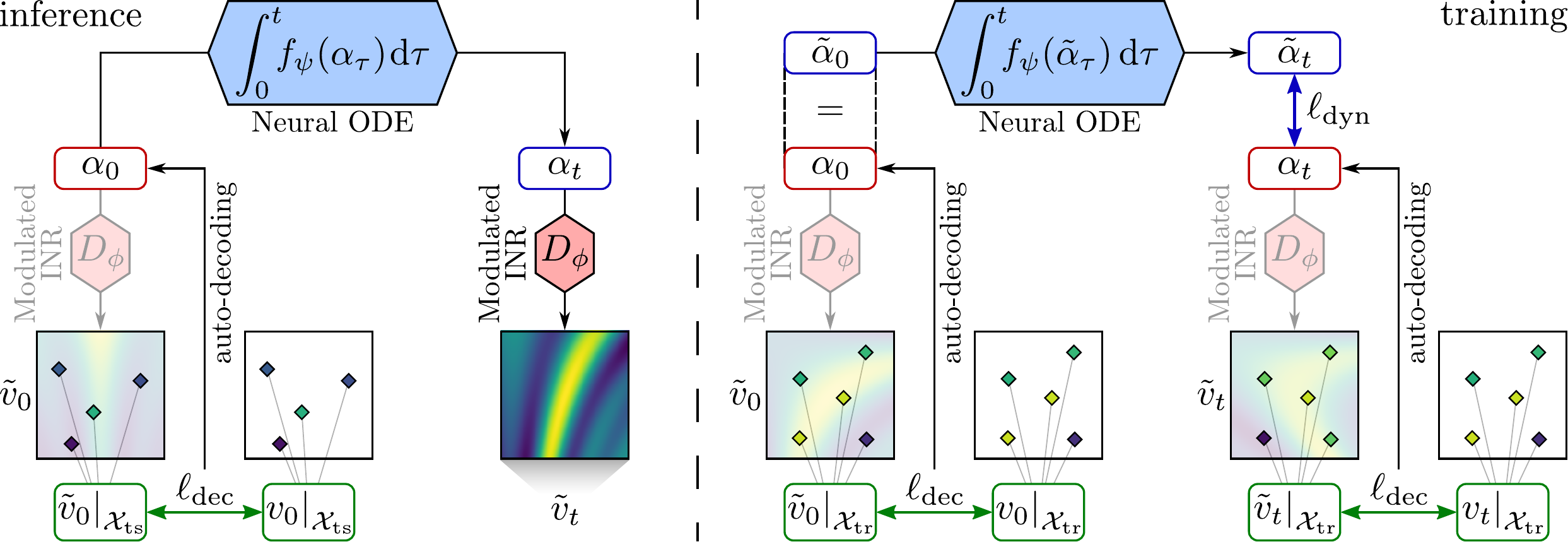}
    \caption{
        \label{fig:model_schema}
        Proposed \ours model.
        Inference (left): given a new initial condition observed on a grid $\gX_{\ts}$, $\rightvert*{v_0}_{\gX_{\ts}}$, forecasting amounts at decoding $\alpha_t$ to $\tilde{v}_t$, by unrolling $\alpha_0$ with a time-continuous ODE dynamics model $f_\psi$.
        Train (right): given an observation grid $\gX_{\tr}$ and a space-continuous decoder $D_\phi$, $\alpha_t$ is learned by auto-decoding \sut $D_\phi(\alpha_t)|_{\gX_{\tr}}=\rightvert*{v_t}_{\gX_{\tr}}$; its evolution is then modeled with $f_\psi$.
    }
    \vspace{-1.25em}
\end{figure}
We present \ours, the first space\,/\,time-continuous model that tackles all prediction tasks of \cref{sec:tasks}, without the above limitations.
We specify \ours's inference procedure (\cref{sec:inference}), illustrated in \cref{fig:model_schema} (left), then introduce each of its components (\cref{sec:components}) and how they are trained (\cref{sec:training}, \cref{fig:model_schema} (right)).
Finally, we detail our implementation based on amplitude modulation, a novel INR parameterization for spatiotemporal data which performs separation of variables (\cref{sec:inr_archi}).

\subsection{Inference model}
\label{sec:inference}
As explained in \cref{sec:tasks}, we aim at estimating the flow $\Phi$ in \Eqref{eq:flow}, so that our model can be trained on an observed grid $\gX_{\tr}$ and perform inference given a new one $\gX_{\ts}$, both possibly irregular.
To this end, we leverage a space- and time-continuous formulation, independent of a given data discretization.
At inference, \ours starts from \rev{a single} initial condition $v_0\in\gV$ and uses a flow to forecast its dynamics.
\ours first embeds spatial observations from $v_0$ into a latent vector $\alpha_0$ of small dimension $d_\alpha$ via an encoder of spatial functions $E_{\varphi}\colon \gV \to \R^{d_\alpha}$ \textsc{(enc)}.
Then, it unrolls a latent time-continuous dynamics model $f_{\psi}\colon \smash{\R^{d_\alpha}\rightarrow\R^{d_\alpha}}$ given this initial condition \textsc{(dyn)}.
Finally, it decodes latent vectors via a decoder $D_\phi\colon \smash{\R^{d_\alpha}}\rightarrow\gV$ into a spatial function \textsc{(dec)}.
At any time $t$, $D_\phi$ takes as input $\alpha_t$ and outputs a function $\tilde{v}_t\colon\Omega \to \R^n$.
This results in the following model, illustrated in \cref{fig:model_schema} (left) and whose components are detailed in \cref{sec:components}:
\begin{align} \label{eq:dyn_sys}
     \textsc{(enc)} ~\alpha_{0}=E_{\varphi}(v_{0}), && \textsc{(dyn)} ~\od{\alpha_t}{t} = \app{f_{\psi}}{\alpha_t}, && \textsc{(dec)} ~\forall t, \tilde{v}_{t} = D_{\phi}(\alpha_{t}).
\end{align}

\subsection{Components}
\label{sec:components}

\paragraph{Encoder: $\alpha_t=E_\varphi(v_t)$.} The encoder computes a latent vector $\alpha_t$ given observation $v_t$ at any time $t$.
It is used in two different contexts, respectively for train and test.
At train time, given an observed trajectory $v_{\gT}=\{v_t\}_{t\in\gT}$, it will encode any $v_t$ into $\alpha_t$ (see \cref{sec:training}).
At inference time, only $v_0$ is available, and then only $\alpha_0$ is computed to be used as initial value for the dynamics.
Given the decoder $D_\phi$, $\alpha_t$ is a solution to the inverse problem $D_\phi(\alpha_t)=v_t$.
We solve this inverse problem with auto-decoding \citep{Park2019}.
Denoting $\ell_{\text{dec}}(\phi,\alpha_{t};v_t)=\|D_\phi(\alpha_t)-v_t\|_2^2$ the decoding loss where $\euclideannorm*{\cdot}_2$ is the euclidean norm of a function and $K$ the number of update steps, auto-decoding defines $E_\varphi$ as:
\begin{align}
    \label{eq:autodec} \textstyle E_\varphi(v_t) = \alpha_t^K, && \text{where }~~\alpha_t^0 = \alpha_t;~~\forall k>0, \alpha_t^{k+1} = \alpha_t^{k} - \eta \nabla_{\alpha_t} \ell_{\text{dec}}(\phi,\alpha^{k}_{t};v_t)~~\text{ and }~~\varphi=\phi.
\end{align}
In practice, we observe a discretization $(\gX_{\tr}, \gX_{\ts})$ and accordingly approximate the norm in $\ell_{\text{dec}}$ as in \Eqref{eq:bi_level}.
Compared to auto-encoding, auto-decoding underfits less \citep{Kim2019} and is more flexible: without requiring specialized encoder architecture, it handles free-formed (irregular or on a manifold) observation grids as long as the decoder shares the same property.

\begin{wrapfigure}[20]{r}{0.19\textwidth}
\vspace{-0.5cm}
    \centering
    \includegraphics[width=0.9\linewidth]{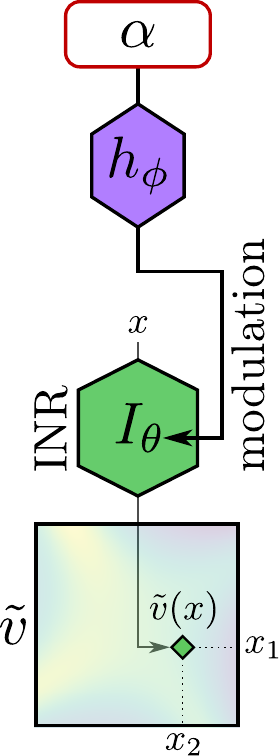}
    \caption{Decoding via INR \Eqref{eq:emission}. \label{fig:dec}}
\end{wrapfigure}
\paragraph{Decoder: $\tilde{v}_t=D_\phi(\alpha_t)$.}
We define a flexible decoder using a coordinate-based INR network with parameters conditioned on $\alpha_t$.
An INR $I_\theta\colon\Omega\to\R^n$ is a space-continuous model parameterized by $\theta\in\R^{d_\theta}$ defined on domain $\Omega$.
It approximates functions independently of the observation grid, \eg it handles irregular grids and changing observation positions unlike FNO and DeepONet.
Thus, it constitutes a flexible alternative to operators suitable to auto-decoding.
To implement the conditioning of the INR's parameters, we use a hypernetwork \citep{Ha2017} $h_{\phi}\colon \R^{d_\alpha}\rightarrow \R^{d_\theta}$, as illustrated in \cref{fig:dec}.
It generates high-dimensional parameters $\theta_t \in \smash{\R^{d_\theta}}$ of the INR given the low-dimensional latent vector $\alpha_t\in\R^{d_\alpha}$.
Hence, the decoder $D_{\phi}$, parameterized by $\phi$, is defined as:
\begin{equation}
    \smash{\forall x\in\Omega, \quad \tilde{v}_t(x)=D_{\phi}(\alpha_{t})(x) \triangleq I_{h_{\phi}(\alpha_{t})}(x).} \label{eq:emission}
\end{equation}
\rev{The decoder's predictions at all spatial locations $x\in\Omega$ thus all depend on $\alpha_t$.} We provide further details on the precise implementation in \Cref{sec:inr_archi}.

\paragraph{Dynamics model: $\od{\alpha_t}{t} = \app{f_{\psi}}{\alpha_t}$.}
Finally, the dynamics model $f_{\psi}\colon \R^{d_\alpha}\rightarrow\R^{d_\alpha}$ defines a flow via an ODE in the latent space.
The initial condition can be defined at any time $t$ by encoding with $E_\varphi$ the corresponding input function $v_t$.

\paragraph{Overall flow.}
Combined altogether, our components define the following flow in the input space that can approximate the data flow $\Phi$ in \Eqref{eq:flow}:
\begin{equation} \label{eq:flow-dino}
    \forall(t, \tau), \qquad (v_t, \tau) \textstyle \mapsto D_\phi\bigl(E_\varphi(v_t) + \int_t^{t+\tau} f_\psi(\alpha_{\tau'}) \dif \tau'\bigr) \quad \text{~where~} \alpha_t=E_\varphi(v_t).
\end{equation}
To summarize, \ours defines a time-continuous latent temporal model with a space-continuous emission function $D_{\phi}$, combining the flexibility of space and time continuity.
This is fully novel to our knowledge, as prior latent approaches are discretized (\cf \cite{Fraccaro2018} for state-space models).

\subsection{Training}
\label{sec:training}
We present the training procedure, illustrated in \cref{fig:model_schema} (right), of the previous components.
We use a two-stage optimization process, close to recent work in video prediction \citep{Yan2021}.
Given the train sequences $\gD$, we first use auto-decoding to obtain the latent vectors $\alpha_{\gT}=\{\alpha_t^{v}\}_{t\in\gT, v\in\gD}$ and the decoder parameters $\phi$.
We then learn the parameters of the dynamics $\psi$ by modeling the latent flow over $\alpha_t^{v}, \forall v \in \gD$.
We detail this procedure in \cref{app:algo}, which can be formalized as a \rev{two-stage} optimization problem that we solve in parallel without inducing training instability (\cf \Cref{app:convergence_analysis}):
\begin{equation} \label{eq:bi_level}
    \begin{aligned}
        \textstyle \min_{\psi} \quad \ell_{\text{dyn}}(\psi,\alpha_{\gT}) & \triangleq \textstyle \mathbb{E}_{v \in \gD, t \in \gT} \bigl\|\alpha^{v}_t - \bigl(\alpha^{v}_{0}+\int_{0}^{t} f_\psi(\alpha^{v}_\tau)\dif\tau\bigr)\bigr\|_2^2 \\
        \textstyle \text{\sut~} \alpha_{\gT},\phi=\argmin_{\alpha_{\gT},\phi} \quad \ell_{\text{dec}}(\phi,\alpha_{\gT}) & \triangleq \textstyle \textstyle \mathbb{E}_{v \in \gD, x \in \gX_{\tr}, t \in \gT} \bigl\|v_{t}(x) - D_{\phi}(\alpha_t^v)(x)\bigr\|_2^2.
    \end{aligned}
\end{equation}

\subsection{Decoder implementation via amplitude-modulated INRs}
\label{sec:inr_archi}
We now specify our implementation of decoder $D_{\phi}$ in \Eqref{eq:emission}.
This includes the definition of the INR architecture $I_\theta$ and of the hypernetwork $h_{\phi}$.
We introduce for the latter a new method called amplitude modulation, which implements a space-time separation of variables.

\paragraph{$I_{\theta}$ as FourierNet.}
We implement $I_{\theta}$ as a FourierNet, a state-of-the-art INR architecture, which instantiates a Multiplicative Filter Network \citep[MFN,][]{Fathony2021}.
A FourierNet relies on the recursion in \Eqref{eq:mfn}, where $x\in\Omega$ is an input spatial location, $z^{(l)}(x)$ is the hidden feature vector at layer $l$ for $x$ and $s_{\omega^{(l)}}(x)=[\cos(\omega^{(l)}x), \sin(\omega^{(l)}x)]$ is a Fourier basis:
\begin{equation} \label{eq:mfn}
    \begin{cases}
        \begin{aligned}
        z^{(0)}(x)&=s_{\omega^{(0)}}(x), \qquad z^{(L)}(x) = W^{(L-1)} z^{(L-1)}(x)+b^{(L-1)},\\
        z^{(l)}(x) &= \bigl(W^{(l-1)} z^{(l-1)}(x)+b^{(l-1)}\bigr) \odot s_{\omega^{(l)}}(x) \quad \text{ for } l\in\lrbrackets*{1, L-1},
        \end{aligned}
    \end{cases}
\end{equation}
where we fix $W^{(0)}=0$, $b^{(0)}=1$, $s_{\omega^{(0)}}(x) = x$ and $\odot$ is the Hadamard product.
Denoting $W=[W^{(l)}]_{l=1}^{L-1}, b=[b^{(l)}]_{l=1}^{L-1}, \omega=[\omega^{(l)}]_{l=1}^{L-1}$, we fit a FourierNet to an input function $v$ observed on a grid $\gX$ by learning $\{W, b, \omega\}$ \sut $\forall x\in\gX, z^{(L)}(x)=v(x)$.
In practice, we observe that fixing $\omega$ uniformly sampled performs similarly as learning them, so we exclude them from training parameters.

FourierNets are interpretable, a property we leverage to  separate time and space via amplitude modulation.
\cite{Fathony2021} show that for some $M\gg L\in\mathbb{N},$ there exist a set of coefficients $\{c_{j}^{(m)}\}_{m=1}^M$ that depend individually on $\{W, b\}$ as well as a set of parameters $\{\gamma^{(m)}\}_{m=1}^M$ that depend individually on those of the filters $\omega$ \sut the $j$\textsuperscript{th} dimension of $z^{(L)}(x)$ can be expressed as:
\begin{equation} \label{eq:mfn_form}
     \textstyle z^{(L)}_j(x) = \sum_{m=1}^{M} c_{j}^{(m)} s_{\gamma^{(m)}}(x)+\text{bias}.
\end{equation}
\Eqref{eq:mfn_form} involves a basis of spatial functions $\{s_{\gamma^{(m)}}\}_{m=1}^M$ evaluated on $x$ and the amplitudes of this basis $\{c_{j}^{(m)}\}_{m=1}^M$.
Note that \Eqref{eq:mfn_form} can be extended to other choices of $s_{\omega^{(l)}}$ \citep{Fathony2021}.

\paragraph{$h$ as amplitude modulation.}
\begin{wrapfigure}[20]{r}{0.29\textwidth}
\vspace{-0.45cm}
    \centering
    \includegraphics[width=\linewidth]{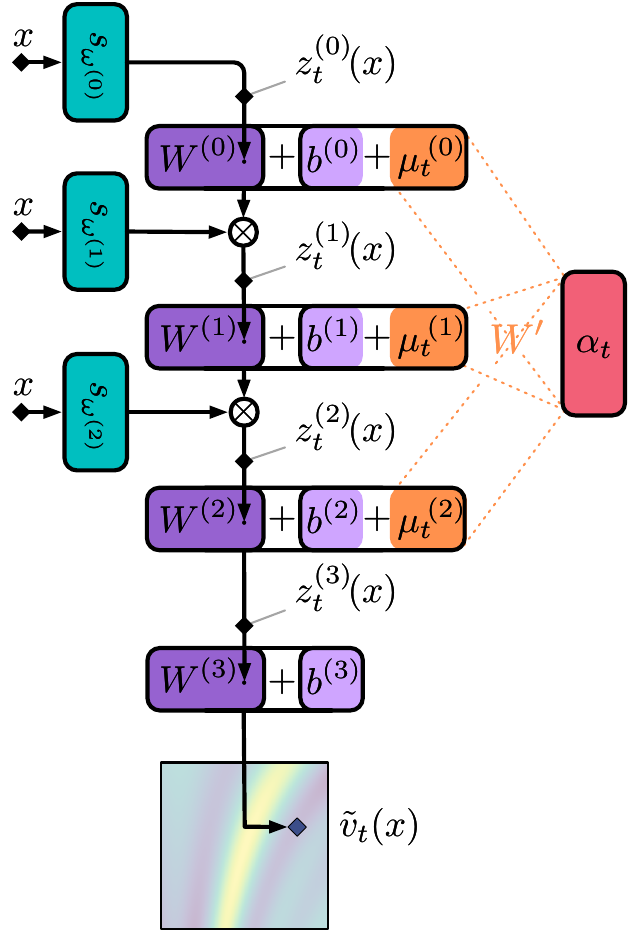}
    \vspace{-0.55cm}
    \caption{Amplitude modulation --~\Eqref{eq:shift_modul}. \rev{$z_t^{(l-1)}$ is input to the $l^{\text{th}}$ linear layer and combined with the spatial basis $s_{\omega^{(l)}}$ via Hadamard product.}} \label{fig:modulation_dino}
\end{wrapfigure}
$h$ generates the INR's parameters $\theta_t$ given $\alpha_t$ to model a target input function $v_t$.
We implement $h$ as elementwise shift and scale transformations \citep[FiLM,][]{Perez2018} of the linear layers parameters $W, b$ (excluding those of the filters $\omega$).
Then, in \Eqref{eq:mfn_form}, amplitudes $c_j^{(m)}$ only depend on time while the basis functions $s_{\gamma^{(m)}}$ only depend on space: this corresponds to \rev{a modeling assumption of separation of variables \citep{LeDret2016} in $v$}.
We call our technique amplitude modulation.
In practice, as \cite{Dupont2022}, we consider latent shift transformations (\cref{fig:modulation_dino}), detailed in \Eqref{eq:shift_modul}.
\Eqref{eq:shift_modul} extends \Eqref{eq:mfn} by introducing a shift term $\mu_t^{(l-1)}$ at each layer $l$, defined as $\mu_t^{(l-1)} = W^{\prime(l-1)} \alpha_{t}$, where $W^\prime=[W^{\prime(l-1)}]_{l=1}^{L-1}$ is another weight matrix:
\begin{equation}
    z_{t}^{(l)}(x) = \bigl(W^{(l-1)} z_t^{(l-1)}(x)+b^{(l-1)} + \mu_t^{(l-1)}\bigr) \odot s_{\omega^{(l)}}(x).\label{eq:shift_modul}
\end{equation}
The INR's parameters are defined as $h_{\phi}(\alpha_{t})=\{W; b + W^\prime \alpha_{t}; \omega\}$ where $\phi=\{W, b, W^\prime\}$ are $h$'s parameters.
Thus, amplitude modulation separates time and space.
We show in \Cref{tab:results_full} that it significantly improves performance, particularly time extrapolation.

    \section{Experiments}
\label{sec:experiments}

We assess the spatiotemporal versatility of \ours, following \Cref{sec:tasks}.
We introduce our experimental setting (\Cref{sec:exp_setting}), which includes a variety of challenging PDE datasets, state-of-the-art baselines and forecasting tasks.
Then, we present and comment the experimental results (\Cref{sec:results}).

\begin{table}[t]
    \caption{
        \label{tab:results}
        \textbf{Space and time \rev{generalization}.}
        Train and test observation grids are equal and subsampled from an uniform 64$\times$64 grid, used for inference.
        We report MSE ($\downarrow$) on the inference time interval $\gT'$, divided within training horizon (\textit{In-t}, $\gT$) and beyond (\textit{Out-t}, outside $\gT$) across subsampling ratios.
    }
    \scriptsize
    \centering
    \sisetup{detect-weight, table-align-uncertainty=true, table-number-alignment=center, mode=text, table-format=1.3e-1,output-exponent-marker = \textsc{e},  exponent-product={}, retain-zero-exponent}
    \scriptsize
    \begin{tabular}{r@{}lS@{\quad~~}SS@{\quad~~}SS@{\quad~~}SS@{\quad~~}S}
        \toprule
        \multicolumn{2}{c}{\multirow{3}{*}[-\dimexpr \aboverulesep + \belowrulesep + \cmidrulewidth]{Model}} & \multicolumn{4}{c}{Navier-Stokes} & \multicolumn{4}{c}{Wave} \\
        \cmidrule(lr){3-6} \cmidrule(lr){7-10}
        & & \multicolumn{2}{c}{Train} & \multicolumn{2}{c}{Test} & \multicolumn{2}{c}{Train} & \multicolumn{2}{c}{Test} \\
        \cmidrule(lr){3-4} \cmidrule(lr){5-6} \cmidrule(lr){7-8} \cmidrule(lr){9-10}
        & & {In-t} & {Out-t} & {In-t} & {Out-t} & {In-t} & {Out-t} & {In-t} & {Out-t} \\
        \midrule
        \multicolumn{10}{c}{$s=5\%$ subsampling ratio} \\[0.3em]
        \ldelim\{{1}{*}[Discrete\,] & I-MP-PDE & 8.154e-03 & 8.166e-03 & \Uline{7.926e-03} & \Uline{8.225e-03} & \Uline{7.055e-04} & \Uline{7.097e-04} & \Uline{1.138e-03} & \Uline{1.116e-03} \\
        \ldelim\{{1}{*}[Operator\,] & DeepONet & \Uline{3.330e-03} & \Uline{7.370e-03} & 1.346e-02 & 1.408e-02 & 8.331e-04 & 9.295e-03 & 1.692e-02 & 3.256e-02 \\
        \ldelim\{{2}{*}[INR\,] & SIREN & 8.741e-3 & 1.767e-1 & 4.303e-2 & 2.126e-1 & 2.738e-3 & 1.818e-2 & 3.339e-2 & 6.964e-2 \\
        & \ours & \B 1.029e-3 & \B 1.655e-3 & \B 1.326e-3 & \B 1.813e-3 & \B 4.088e-5 & \B 4.121e-5 & \B 6.415e-5 & \B 7.392e-5\\
        \midrule
        \multicolumn{10}{c}{$s=25\%$ subsampling ratio} \\[0.3em]
        \ldelim\{{1}{*}[Discrete\,] & I-MP-PDE & \Uline{3.135e-04} & \Uline{7.245e-04} & \Uline{3.476e-04} & \Uline{7.658e-04} & \Uline{3.293e-05} & \Uline{1.108e-04} & \Uline{5.142e-05} & \Uline{1.545e-04} \\
        \ldelim\{{1}{*}[Operator\,] & DeepONet & 9.016e-04 & 5.936e-03 & 9.376e-03 & 1.328e-02 & 5.722e-04 & 1.061e-02 & 1.757e-02 & 3.221e-02\\
        \ldelim\{{2}{*}[INR\,] & SIREN & 5.180e-03 & 2.175e-01 & 2.436e-01 & 3.861e-01 & 8.995e-04 & 1.292e-02 & 1.783e-02 & 5.143e-02\\
        & \ours & \B 1.020e-04 & \B 4.504e-04 & \B 2.646e-04 & \B 5.951e-04 & \B 3.949e-06 & \B 4.436e-06 & \B 1.089e-05 & \B 1.174e-05 \\
        \midrule
        \multicolumn{10}{c}{$s=100\%$ subsampling ratio} \\[0.3em]
        \ldelim\{{2}{*}[Discrete\,] & CNODE & 2.319e-2 & 9.652e-2 & 2.305e-2 & 1.143e-1 & 2.337e-05 & 5.280e-04 & 3.057e-05 & 7.288e-04 \\
        & MP-PDE & 1.140e-04 & \Uline{5.500e-4} & \B 1.785e-04 & \Uline{5.856e-04} & \B 1.718e-07 & \Uline{1.993e-05} & \B 9.256e-07 & \Uline{4.261e-5} \\
        \ldelim\{{2}{*}[Operator\,] & MNO & \B 3.190e-5 & 8.678e-4 & 2.763e-4 & 8.946e-4 & 9.381e-6 & 4.890e-3 & 1.993e-4 & 6.128e-3 \\
        & DeepONet & 1.375e-03 & 6.573e-03 & 9.704e-03 & 1.244e-02 & 6.431e-04 & 1.293e-02 & 1.847e-02 & 3.317e-02 \\
        \ldelim\{{4}{*}[INR\,] & SIREN & 1.066e-3 & 4.336e-1 & 3.874e-1  & 1.037 & 3.674e-4 & 9.956e-3 & 3.013e-2 & 7.842e-2 \\
        & MFN & 1.651e-3 & 1.037e-0 & 2.106e-01 & 1.059e+00 & 1.408e-4 & 1.763e-1 & 4.735e-3 & 2.274e-1   \\
        & \ours (no sep.) & 3.235e-04 & 1.593e-03 & 7.850e-04 & 1.889e-03 & \Uline{2.641e-6} & 4.081e-5 & 5.977e-5 & 2.979e-4 \\
        & \ours & \Uline{8.339e-5} & \B 3.115e-4 & \Uline{2.092e-4} & \B 4.311e-4 & 3.309e-6 & \B 3.506e-6 & \Uline{9.495e-6} & \B 9.946e-6 \\
        \bottomrule
    \end{tabular}
    \vspace{-1em}
\end{table}

\subsection{Experimental setting}
\label{sec:exp_setting}
\paragraph{Datasets.}
We consider the following PDEs defined over a spatial domain $\Omega$, with further details in \cref{app:data}.
\begin{itemize*}
    \item \textbf{2D Wave equation} (\textit{Wave}) is a second-order PDE $\pd[2]{u}{t} = c^2\Delta u$.
    $u$ is the displacement \wrt the rest position and $c$ is the wave traveling speed.
    We consider its first-order form, so that $v_t = (u_t, \pd{u_t}{t})$ has a two-dimensional output ($n=2$).
    \item \textbf{2D Navier Stokes} (\textit{Navier-Stokes}, \citealp{Stokes1851}) corresponds to an incompressible fluid dynamics $\od{v}{t} = -u \nabla v + \nu \Delta v + f, v = \nabla \times u, \nabla u = 0$, where $u$ is the velocity field and $v$ the vorticity.
    $\nu$ is the viscosity and $f$ is a constant forcing term; $n=1$.
    \item \textbf{3D Spherical shallow water} (\textit{Shallow-Water}, \citealp{Galewsky2004}): it involves the vorticity $w$, tangent to the sphere's surface, and the thickness of the fluid $h$.
    The input is $v_t = (w_t, h_t)$; $n=2$.
\end{itemize*}

\paragraph{Baselines.}
We reimplement representative models from \Cref{sec:related_work,tab:comparison} and adapt them to our multi-dimensional datasets.
\begin{itemize*}
    \item \textbf{CNODE} \citep{Ayed2020} combines a CNN and an ODE solver to handle regular grids.
    \item \textbf{MP-PDE} \citep{Brandstetter2022} uses a GNN to handle free-formed grids, yet is unable to predict outside the observation grid.
    We developed an interpolative extension, \textbf{I-MP-PDE}, to handle this limitation; it performs bicubic interpolation on the observed grid and training is done on the resulting interpolation.
    \item \textbf{MNO} \citep{Li2021b} is an autoregressive version of FNO \citep{Li2021} for regular grids; it can be evaluated on new uniform grids.
    \item \textbf{DeepONet} \citep{Lu2021}, considered autoregressively \citep{Wang2021} where we remove time from the trunk net's input, can be evaluated on new spatial locations without interpolation.
    \item \textbf{SIREN} \citep{Sitzmann2019} and \textbf{MFN} \citep{Fathony2021} are two INR methods which we extend to fit our setting.
    We consider an agnostic setting, \ie without the knowledge of the differential equation, and perform sequence conditioning to generalize to more than a trajectory.
    This is achieved by learning a latent vector with auto-decoding; it is then concatenated to the spatial coordinates.
\end{itemize*}

\paragraph{Tasks.}
We evaluate models on various forecasting tasks which combine the evaluation scenarios of \cref{sec:tasks}.
Performance is measured by the prediction Mean Squared Error (MSE) given only an initial condition.
\begin{itemize*}
    \item \textbf{Space and time \rev{generalization}.}
    We consider a uniform grid $\gX'$ for inference.
    Training is performed on different observations grids $\gX_{\tr}$ subsampled from $\gX'$ with different ratios, $s\in\{5\%, 25\%, 50\%, 100\%\}$ where $s=100\%$ corresponds to the full inference grid, \ie $\gX_{\tr} = \gX'$.
    In this setting, we consider that all trajectories (\textit{train} and \textit{test}) share the same observation grid $\gX_{\tr} = \gX_{\ts}$.
    We evaluate MSE error on $\gX'$ over the train time interval (\textit{In-t}) and beyond (\textit{Out-t}) at each subsampling ratio.
    \item \textbf{Flexibility \wrt input grid.}
    We vary the test observation grid, \ie $\gX_{\ts}\neq\gX_{\tr}$ and perform inference on $\gX'=\gX_{\ts}$, \ie on the test observation grid (\textit{In-s}) under two settings:
    \begin{itemize*}[label=$\triangleright$]
        \item \textbf{Generalizing across grids:} $\gX_{\tr}, \gX_{\ts}$ are subsampled differently from the same uniform grid; $s_{\tr}$ (\resp$s_{\ts}$) is the train (\resp test) subsampling ratio.
        \item \textbf{Generalizing across resolutions:} $\gX_{\tr}, \gX_{\ts}$ are subsampled with the same ratio $s$ from two uniform grids with different resolutions; the train resolution is fixed to $r_{\tr}=64$ while we vary the test resolution $r_{\ts}\in\{32, 64, 256\}$.
    \end{itemize*}
    \item \textbf{Data on manifold.}
    We consider a PDE on a sphere and combine several evaluation scenarios, as described later.
    \item \textbf{Finer time resolution.}
    We consider an inference time grid $\gT'$ with a finer resolution than the train one $\gT$.
\end{itemize*}

\subsection{Results}
\label{sec:results}

\begin{table}[t]
    \caption{\label{tab:results_flex}\textbf{Flexibility \wrt input grid.} Observed test / train grid differ ($\gX_{\ts}\neq\gX_{\tr}$). We report \textit{test} MSE ($\downarrow$) for \textit{Navier-Stokes} on $\gX'=\gX_{\ts}$ (\textit{In-s}). \cgreen{Green}\cyellow{Yellow}\cred{Red} mean excellent, good, poor MSE.}
    \centering
    \subfloat[\label{tab:results_new_grid_2} \textbf{Generalization across grids}: $\gX_{\tr}, \gX_{\ts}$ are subsampled with different ratios $s_{\tr}\neq s_{\ts}$ among $\{5, 50, 100\}$\% from the same uniform 64$\times$64 grid.]{
        \centering
        \sisetup{detect-weight, table-align-uncertainty=true, table-number-alignment=center, output-exponent-marker = \textsc{e}, table-format=1.3e-1,  exponent-product={}, retain-zero-exponent, mode=text}
        \scriptsize
        \begin{tabular}{llSSSSSSSSSS}
             \toprule
             Subsampling & Test$\rightarrow$ & \multicolumn{2}{c}{$s_{\ts}=5\%$} & \multicolumn{2}{c}{$s_{\ts}=50\%$} & \multicolumn{2}{c}{$s_{\ts}=100\%$} \\
              \cmidrule(lr){3-4} \cmidrule(lr){5-6} \cmidrule(lr){7-8}
             Train $\downarrow$ & & {In-t} & {Out-t} & {In-t} & {Out-t} & {In-t} & {Out-t} \\
             \midrule
              \multirow{2}{*}{$s_{\tr}=5\%$} & MP-PDE & \cellcolor{cred} 1.330e-01 & \cellcolor{cred} 3.852e-01 & \cellcolor{cred} 1.859e-01 & \cellcolor{cred} 6.680e-01 & \cellcolor{cred} 2.105e-01 & \cellcolor{cred} 7.120e-01 \\
              & \ours{} & \cellcolor{cyellow} \B 1.494e-03 & \cellcolor{cyellow}\B 2.291e-03 & \cellcolor{cyellow} \B 1.257e-03 & \cellcolor{cyellow} \B 1.883e-03 & \cellcolor{cyellow} \B 1.287e-03 & \cellcolor{cyellow} \B 1.947e-03 \\
             \midrule
             \multirow{2}{*}{$s_{\tr}=50\%$} & MP-PDE & \cellcolor{cred} 4.494e-02 & \cellcolor{cred} 9.403e-02 & \cellcolor{cyellow} 4.793e-03 & \cellcolor{cred} 1.997e-02 & \cellcolor{cyellow} 6.330e-03 & \cellcolor{cred} 3.712e-02  \\
             & \ours & \cellcolor{cgreen} \B 2.470e-04 & \B \cellcolor{cgreen} 4.697e-04 & \cellcolor{cgreen} \B 2.073e-04 & \cellcolor{cgreen} \B 4.284e-04 & \cellcolor{cgreen} \B 2.058e-04 & \cellcolor{cgreen} \B 4.361e-04 \\
            \midrule
            \multirow{2}{*}{$s_{\tr}=100\%$} & MP-PDE & \cellcolor{cred} 1.358e-01 & \cellcolor{cred} 3.355e-01 & \cellcolor{cred} 1.182e-02 & \cellcolor{cred} 2.664e-02 & \B \cellcolor{cgreen} 1.785e-04 & \cellcolor{cgreen} 5.856e-04 \\
             & \ours & \B \cellcolor{cgreen} 2.495e-04 & \B \cellcolor{cgreen} 4.805e-04 & \B \cellcolor{cgreen} 2.109e-04 & \cellcolor{cgreen} \B 4.325e-04 & \cellcolor{cgreen} 2.092e-4 & \cellcolor{cgreen} \B 4.311e-4 \\
             \bottomrule
        \end{tabular}}

    \vspace{-0.5em}
    \subfloat[\label{tab:results_new_grid} \textbf{Generalization across resolutions}: $\gX_{\ts}$ (resp.\ $\gX_{\tr}$) are subsampled at the same ratio $s\in\{5, 100\}$\% from different uniform grids with resolution $r_{\ts}\in\{32, 64, 256\}$ (resp.\ $r_{\tr}=64$).]{
        \centering
        \sisetup{detect-weight, table-align-uncertainty=true, table-number-alignment=center, output-exponent-marker = \textsc{e}, table-format=1.3e-1,  exponent-product={}, retain-zero-exponent, mode=text}
        \scriptsize
        \begin{tabular}{llSSSSSS}
             \toprule
             \multicolumn{2}{r}{Test resolution $\rightarrow$} & \multicolumn{2}{c}{$r_{\ts}=32$ - $\gX_{\ts}\neq\gX_{\tr}$} & \multicolumn{2}{c}{$r_{\ts}=64$ - $\gX_{\ts}=\gX_{\tr}$} & \multicolumn{2}{c}{$r_{\ts}=256$ - $\gX_{\ts}\neq\gX_{\tr}$} \\
             \cmidrule(lr){3-4} \cmidrule(lr){5-6} \cmidrule(lr){7-8}
             Subsampling $\downarrow$ & & {In-t} & {Out-t} & {In-t} & {Out-t} & {In-t} & {Out-t} \\
             \midrule
             \multirow{2}{*}{$s=5\%$} & MP-PDE & \cellcolor{cred} 3.209e-01 & \cellcolor{cred} 6.472e-01 & \cellcolor{cgreen} \B 2.465e-04 & \cellcolor{cyellow} 1.105e-03 & \cellcolor{cred} 2.239e-01 & \cellcolor{cred} 8.253e-01 \\
              & \ours{} & \cellcolor{cyellow} \B 5.308e-03 & \cellcolor{cyellow} \B 9.544e-03 & \cellcolor{cgreen} 2.533e-04 & \cellcolor{cgreen} \B 8.832e-04 & \cellcolor{cyellow} \B 1.991e-03 & \cellcolor{cyellow} \B 2.942e-03 \\
             \cmidrule(lr){1-8}
             \multirow{3}{*}{$s=100\%$} & MNO & \cellcolor{cyellow} 4.547e-03 & \cellcolor{cyellow} 9.281e-03 & \cellcolor{cgreen} \B 1.277e-04 & \cellcolor{cgreen} 8.525e-04 & \cellcolor{cyellow} 2.174e-3 & \cellcolor{cyellow} 4.975e-03 \\
             & MP-PDE & \cellcolor{cred} 4.194e-02 & \cellcolor{cred} 9.109e-02 & \cellcolor{cgreen} 1.597e-04 & \cellcolor{cgreen} 6.483e-04 & \cellcolor{cred} 4.648e-02 & \cellcolor{cred} 1.381e-01 \\
             & \ours{} & \cellcolor{cgreen} \B 2.321e-04 & \cellcolor{cgreen} \B 6.386e-04 & \cellcolor{cgreen} 2.320e-04 & \cellcolor{cgreen} \B 6.385e-04 & \cellcolor{cgreen} \B 2.322e-04 & \cellcolor{cgreen} \B 6.385e-04 \\
             \bottomrule
        \end{tabular}}
        \vspace{-2em}
\end{table}

\paragraph{Space and time \rev{generalization}.}
We report prediction MSE in \Cref{tab:results} for varying subsampling ratios $s\in\{5\%, 25\%, 100\%\}$ on \textit{Navier-Stokes} and \textit{Wave}.
\Cref{app:full_results} provides a fine-grained evaluation inside the train observation grid (\textit{In-s}) or outside (\textit{Out-s}) and additionally reports the results for $s=50\%$.
We visualize some predictions in \Cref{app:prediction}.
\ours is compared to all baselines when $s=100\%$, \ie $\gX'=\gX_{\tr}=\gX_{\ts}$, and otherwise it is compared only to models which handle irregular grids and prediction at arbitrary spatial locations (DeepONet, SIREN, MFN, I-MP-PDE).
\begin{itemize*}
    \item \textbf{General analysis.}
    We observe that all models degrade when the subsampling ratio $s$ decreases.
    \ours performs competitively overall: it achieves the best \textit{Out-t} performance on all subsampling settings, it outperforms all the baselines on low subsampling ratios and performs comparably to the competitive discretized (MP-PDE, CNODE) and operator (MNO) alternatives when $s=100\%$, \ie when observation and inference grids are equal.
    Note that this fully observed setting is favorable for CNODE, MP-PDE and MNO, designed to perform inference on the observation grid.
    This can be seen in \cref{tab:results}, where \ours is  slightly outperformed only for few settings.
    MP-PDE is significantly better only on \textit{Wave} for \textit{In-t}.
    Overall, CNNs and GNNs exhibit good performance for spatially local dynamics like \textit{Wave}, while INRs (like \ours) and MNO are more adapted to global dynamics like \textit{Navier-Stokes}.
    \item \textbf{Analysis per model.}
    MP-PDE is the most competitive baseline across datasets as it combines a strong and flexible encoder (GNNs) to a good dynamics model; however, it cannot predict outside the observation grid (\emph{Out-s}).
    To keep a strong competitor, we extend this baseline into its interpolative version I-MP-PDE on subsampled settings.
    I-MP-PDE is competitive for high subsampling ratios, \eg $s\in\{50\%,100\%\}$ but underperforms \wrt \ours at lower subsampling ratios due to the accumulated interpolation error.
    MNO is a competitive baseline on \textit{Navier-Stokes}, performing on par with MP-PDE and \ours inside the training horizon (\textit{In-t}); its performance on \textit{Out-t} degrades more significantly compared to other models, especially \ours.
    DeepONet is more flexible than MP-PDE as it can predict at arbitrary locations.
    As no interpolation error is introduced, it outperforms I-MP-PDE for $s=5\%$ on \textit{train} data.
    Yet, we observe that it underperforms especially on \textit{Out-t} \wrt its alternatives.
    Finally, we observe that SIREN and MFN fit correctly the train horizon \textit{In-t} on \textit{train}, yet generalize poorly outside this horizon \textit{Out-t} or on new initial conditions (\textit{test}).
    This is in accordance with our analysis of \cref{sec:related_work}; we highlight that this is not the case for \ours which extrapolates temporally and generalizes to new initial conditions thanks to its sequential modeling of the flow.
    Thus, \textit{\ours is currently the state-of-the-art INR model for spatiotemporal data}.
    \item \textbf{Modulation.}
    We observe that modulating both amplitudes and frequencies (row ``\ours (no sep.)'' in \Cref{tab:results}) degrades performance \wrt \ours (row ``\ours'' in \Cref{tab:results}) that only modulates amplitudes.
    Amplitude modulation enables long temporal extrapolation and reduces the number of parameters.
    Hence, as opposed to \ours (no sep.) which is outperformed by some baselines, time-space variable separation in \ours is an essential ingredient of the model to reach state-of-the-art levels.
\end{itemize*}

\paragraph{Flexibility \wrt input grid.}  \begin{wrapfigure}[23]{hR}{0.375\linewidth}
\vspace{-0.95cm}
    \begin{minipage}{\linewidth}
        \centering
        \includegraphics[width=\linewidth]{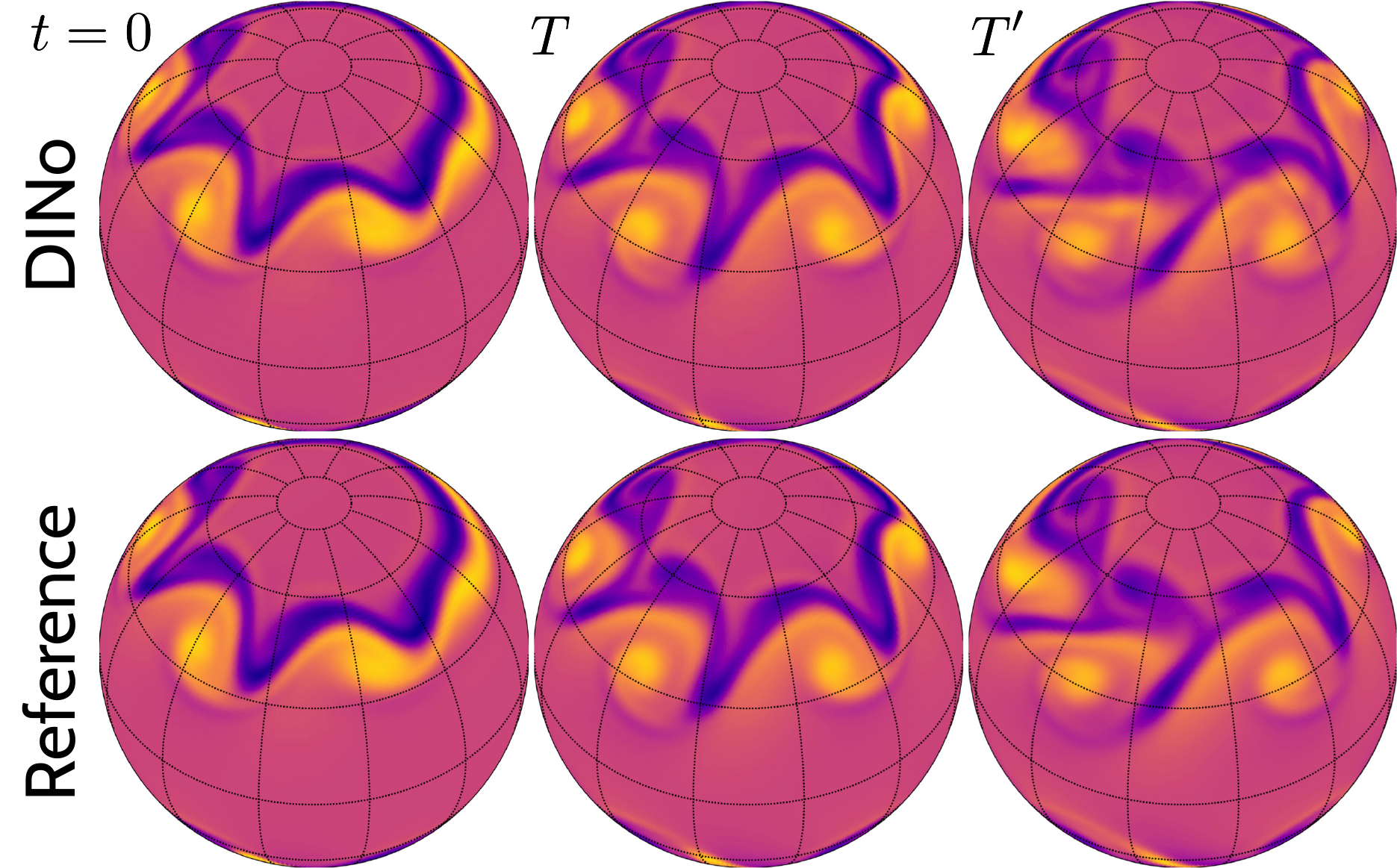} \\
        \vspace{0.05cm}
        \sisetup{detect-weight, table-align-uncertainty=true, table-number-alignment=center, output-exponent-marker = \textsc{e}, table-format=1.3e-1,  exponent-product={}, retain-zero-exponent, mode=text}
        \footnotesize
            \begin{tabular}{lSS}
                \toprule
                Model & {In-t} & {Out-t} \\
                \midrule
                I-MP-PDE & 1.908e-03 & 7.240e-03 \\
                \ours & \B 1.063e-04 & \B 6.466e-04 \\
                \bottomrule
            \end{tabular}
    \end{minipage}
        \vspace{-0.15cm}
    \captionof{figure}{\textbf{Data on manifold.} \ours's \textit{Shallow-Water} superresolution \textit{test} prediction (top) against the reference (middle), with \textit{test} MSE ($\downarrow$) (bottom). \label{fig:results_sphere}}

    \captionof{table}{\textbf{Finer time resolution.} \textit{Test} MSE ($\downarrow$) under $\gT'$ for \textit{Navier-Stokes}.\label{tab:temp_superresol}}
    \sisetup{detect-weight, table-align-uncertainty=true, table-number-alignment=center, output-exponent-marker = \textsc{e}, table-format=1.3e-1,  exponent-product={}, retain-zero-exponent, mode=text}
    \vspace{0.3\baselineskip}
        \scriptsize
        \centering
        \begin{tabular}{lSS}
             \toprule
             Model & {In-t} & {Out-t} \\
             \midrule
             I-\ours (linear) & 3.459e-04 & 5.598e-04 \\
             I-\ours (quadratic) & 2.165e-04 & 4.473e-04 \\
             \ours (ODE solve) & \B 2.151e-04 & \B 4.388e-04 \\
             \bottomrule
        \end{tabular}
\end{wrapfigure}

We consider in \cref{tab:results_flex} \textit{Navier-Stokes} and compare \ours to the most competitive baselines, MP-PDE and MNO (with $s=100\%$ subsampling ratio).
\begin{itemize*}
    \item \textbf{Generalizing across grids.}
    In \Cref{tab:results_new_grid_2}, we consider that the test observation grid $\gX_{\ts}$ is different from the train one $\gX_{\tr}$. This occurs when sensors differ between two observed trajectories. We vary the subsampling ratio for the train observation grid $s_{\tr}$ and the test one $s_{\ts}$. We report \textit{test} MSE on new grids $\gX'=\gX_{\ts}$.
    We observe that \ours is very robust to changing grids between \textit{train} and \textit{test}, while MP-PDE's performance degrades, especially for low subsampling ratios, \eg 5\%. For reference, we report in \Cref{tab:results_new_grid_2_full} \Cref{app:full_results} (result col.~1) the performance when $\gX'=\gX_{\tr}$, where MP-PDE is substantially better.
    \item \textbf{Generalizing across spatial resolutions.}
    In \Cref{tab:results_new_grid} we vary the test resolution $r_{\ts}$. We train at a resolution $r_{\tr}=64$ and perform inference at resolutions $r_{\ts}\in\{32, 64, 256\}$.
    For that, we build a high-fidelity 256$\times$256 simulation dataset and downscale it to obtain the other resolutions.
    We observe that \ours's performance is the stablest across resolutions in the uniform or irregular setting.
    MNO is also relatively stable but is only applicable to uniform grids while MP-PDE is particularly brittle, especially for a $5\%$ ratio.
\end{itemize*}

\paragraph{Data on manifold.}
We consider in \Cref{fig:results_sphere} \textit{Shallow-Water} in a super-resolution setting: test resolution is twice the train one, close to weather prediction applications.
We observe an irregular 3D Euclidean coordinate grid $\smash{\gX_{\tr} = \gX_{\ts} \subset\R^3}$ shared for \textit{train} and \textit{test}.
It uniformly samples Euclidean positions on the sphere, via the quasi-uniform skipped latitude-longitude grid \citep{Weller2012}.
We predict the PDE on \textit{test} trajectories with a conventional latitude-longitude inference grid $\gX'$.
At Earth scale, $\gX_{\tr}$ corresponds to a resolution of about \SI{300}{\km}, and $\X'$ to \SI{150}{\km}.
\ours significantly outperforms I-MP-PDE, making it a viable candidate for this complex setting.

\paragraph{Finer time resolution.}
We consider in \Cref{tab:temp_superresol} a longer and ten times finer test temporal grid $\gT'$ than the train grid $\gT$ on \textit{Navier-Stokes}.
We observe the same spatial uniform grid across \textit{train} and \textit{test} and perform inference on this grid.
We compare \ours that performs prediction with an ODE solver, to interpolating coarser predictions obtained at the train resolution (I-\ours).
We report the corresponding \textit{test} MSE.
We observe that the ODE solver accurately extrapolates outside the train temporal grid, outperforming interpolation.
This confirms that \ours benefits from its continuous-time modeling of the flow, providing consistency and stability across temporal resolutions.

    \section{Conclusion}
We propose \ours, a novel space- and time-continuous data-driven PDE forecaster.
\ours handles free-form spatiotemporal conditions encountered in many applications, where existing methods fail.
\ours outperforms recent PDE forecasters on a variety of PDEs and spatiotemporal \rev{generalization} settings, including evaluation on unseen sparse irregular meshes and resolutions.
There are many promising future work such as scaling \ours to real-world problems, \eg weather forecasting, or incorporating recent strategies to adapt to changing dynamics \citep{Kirchmeyer2022}.

\subsubsection*{Acknowledgements}
We thank Emmanuel de Bézenac and Jérémie Donà for helpful insights and discussions on this project.
We also acknowledge financial support from DL4CLIM (ANR-19-CHIA-0018-01) and DEEPNUM (ANR-21-CE23-0017-02) ANR projects.
This study has been conducted using E.U.\ Copernicus Marine Service Information.

    \section*{Reproducibility statement}
We present in \Cref{sec:exp_setting} our experimental setting with datasets, baselines and forecasting tasks. 
The train and test settings are detailed in \Cref{app:data}, including more information on the chosen physical PDE systems. 
We describe \ours's pseudo-code in \Cref{alg:pdes} and provide implementation and hyperparameters details in \Cref{app:implem}.
We provide our source code at \url{https://github.com/mkirchmeyer/DINo}. 
    \bibliography{refs}
    \bibliographystyle{iclr2023_conference}

    \newpage
    \appendix
    \section{Full results}
\label{app:full_results}
We provide in \Cref{tab:results_full} a more detailed version of \Cref{tab:results} for the space-time extrapolation problem where we report the performance \textit{In-s} (on the observation grid) and \textit{Out-s} (outside). We add $s=50$\%.

Then, we report in \Cref{tab:results_new_grid_2_full} a more detailed version of \Cref{tab:results_new_grid_2}, which includes the results of $\gX_{\ts}=\gX_{\tr}$.
This corresponds to our generalization across grids problem.

\begin{table}[H]
    \caption{
        \label{tab:results_full}
        \textbf{Space and time \rev{generalization}.}
        The train and test observation grids are equal; they are subsampled with a ratio $s$ from an uniform 64$\times$64 grid fixed here to be the inference grid $\gX'$.
        We report MSE ($\downarrow$) on $\gX'$ (on the observation grid \textit{In-s}, outside \textit{Out-s} or on both \textit{Full}) and the inference time interval $\gT'$, divided within training horizon (\textit{In-t}, $\gT$) and beyond (\textit{Out-t}, outside $\gT$) across subsampling ratios $s\in\{5\%, 25\%, 50\%, 100\%\}$. Best in \textbf{bold} and second best \underline{underlined}.
    }
    \scriptsize
    \centering
    \sisetup{detect-weight, table-align-uncertainty=true, table-number-alignment=center, mode=text, table-format=1.3e-1,output-exponent-marker = \textsc{e}, exponent-product={},
    retain-zero-exponent}
    \begin{tabular}{llSSSSSSSS}
        \toprule
        & \multirow{3}{*}[-\dimexpr \aboverulesep + \belowrulesep + \cmidrulewidth]{Model} & \multicolumn{4}{c}{Navier-Stokes} & \multicolumn{4}{c}{Wave} \\
        \cmidrule(lr){3-6} \cmidrule(lr){7-10}
        & & \multicolumn{2}{c}{Train} & \multicolumn{2}{c}{Test} & \multicolumn{2}{c}{Train} & \multicolumn{2}{c}{Test} \\
        \cmidrule(lr){3-4} \cmidrule(lr){5-6} \cmidrule(lr){7-8} \cmidrule(lr){9-10}
        & & {In-t} & {Out-t} & {In-t} & {Out-t} & {In-t} & {Out-t} & {In-t} & {Out-t} \\
        \midrule
        \multicolumn{10}{c}{$s=5\%$ subsampling} \\
        \midrule
        \multirow{4}{*}{\begin{turn}{90} {In-s} \end{turn}}
        & I-MP-PDE & \B 3.525e-05 & \Uline{1.295e-03} & \Uline{4.554e-04} & \Uline{1.414e-03} & \B 1.824e-06 & \Uline{8.672e-05} & \Uline{1.113e-05} & \Uline{1.987e-04} \\
        & DeepONet & 4.778e-04 & 4.517e-03 & 1.060e-02 & 1.059e-02 & 2.546e-04 & 8.831e-03 & 1.501e-02 & 3.196e-02 \\
        & SIREN & 5.966e-3 & 1.769e-1 & 4.082e-2 & 2.150e-1 & 1.690e-3 & 1.707e-2 & 2.951e-2 & 6.955e-2 \\
        & \ours & \Uline{1.016e-04} & \B 6.945e-4 & \B 3.623e-04 & \B 8.306e-04 & \Uline{2.250e-06} & \B 5.283e-6 & \B 7.530e-06 & \B 2.146e-5 \\
        \midrule
        \multirow{4}{*}{\begin{turn}{90} Out-s \end{turn}} & I-MP-PDE & 8.550e-03 & 8.515e-03 & \Uline{8.306e-03} & \Uline{8.571e-03} & \Uline{7.412e-04} & \Uline{7.414e-04} & \Uline{1.195e-03} & \Uline{1.163e-03} \\
        & DeepONet & \Uline{3.475e-03} & \Uline{7.515e-03} & 1.361e-02 & 1.426e-02 & 8.624e-04 & 9.318e-03 & 1.702e-02 & 3.259e-02 \\
        & SIREN & 8.882e-3 & 1.767e-1 &  4.314e-2 & 2.124e-1 & 2.791e-3 & 1.823e-2 & 3.359e-2 & 6.965e-2 \\
        & \ours & \B 1.076e-3 & \B 1.704e-3 & \B 1.375e-3 & \B 1.863e-3 & \B 4.285e-5 & \B 4.304e-5 & \B 6.703e-5 & \B 7.659e-5 \\
        \midrule
        \multirow{4}{*}{\begin{turn}{90} Full \end{turn}} & I-MP-PDE & 8.154e-03 & 8.166e-03 & \Uline{7.926e-03} & \Uline{8.225e-03} & \Uline{7.055e-04} & \Uline{7.097e-04} & \Uline{1.138e-03} & \Uline{1.116e-03} \\
        & DeepONet & \Uline{3.330e-03} & \Uline{7.370e-03} & 1.346e-02 & 1.408e-02 & 8.331e-04 & 9.295e-03 & 1.692e-02 & 3.256e-02 \\
        & SIREN & 8.741e-3 & 1.767e-1 & 4.303e-2 & 2.126e-1 & 2.738e-3 & 1.818e-2 & 3.339e-2 & 6.964e-2 \\
        & \ours & \B 1.029e-3 & \B 1.655e-3 & \B 1.326e-3 & \B 1.813e-3 & \B 4.088e-5 & \B 4.121e-5 & \B 6.415e-5 & \B 7.392e-5\\
        \midrule
        \multicolumn{10}{c}{$s=25\%$ subsampling} \\
        \midrule
        \multirow{4}{*}{\begin{turn}{90} In-s \end{turn}}
        & I-MP-PDE &  \Uline{1.447e-04} & \Uline{5.677e-04} & \B 1.763e-04 & \Uline{6.147e-04} & \B 6.754e-07 & \Uline{8.251e-05} & \B 9.253e-07 & \Uline{1.227e-04} \\
        & DeepONet & 7.500e-04 & 5.779e-03 & 9.227e-03 & 1.300e-02 & 5.196e-04 & 1.058e-02 & 1.743e-02 & 3.246e-02 \\
        & SIREN & 4.786e-03 & 2.178e-01 & 2.461e-01  & 3.884e-01 & 8.478e-04 & 1.282e-02 & 1.733e-02 & 5.104e-02\\
        & \ours & \B 8.295e-05 & \B 4.273e-04 & \Uline{2.444e-04} & \B 5.735e-04 & \Uline{3.194e-06} & \B 3.747e-06 & \Uline{8.907e-06} & \B 1.029e-05 \\
        \midrule
        \multirow{4}{*}{\begin{turn}{90} Out-s \end{turn}} & I-MP-PDE & \Uline{3.678e-04} & \Uline{7.748e-04} & \Uline{4.026e-04} & \Uline{8.143e-04} & \Uline{4.330e-05} & \Uline{1.200e-04} & \Uline{6.764e-05} & \Uline{1.648e-04} \\
        & DeepONet & 9.503e-04 & 5.987e-03 & 9.423e-03 & 1.337e-02 & 5.891e-04 & 1.062e-02 & 1.762e-02 & 3.213e-02 \\
        & SIREN & 5.305e-03 & 2.173e-01 & 2.428e-01 & 3.853e-01 & 9.159e-04 & 1.295e-02 & 1.798e-02 & 5.156e-02 \\
        & \ours & \B 1.081e-04 & \B 4.578e-04 & \B 2.711e-04 & \B 6.021e-04 & \B 4.192e-06 & \B 4.657e-06 & \B 1.153e-05 & \B 1.220e-05 \\
        \midrule
        \multirow{4}{*}{\begin{turn}{90} Full \end{turn}} & I-MP-PDE & \Uline{3.135e-04} & \Uline{7.245e-04} & \Uline{3.476e-04} & \Uline{7.658e-04} & \Uline{3.293e-05} & \Uline{1.108e-04} & \Uline{5.142e-05} & \Uline{1.545e-04} \\
        & DeepONet & 9.016e-04 & 5.936e-03 & 9.376e-03 & 1.328e-02 & 5.722e-04 & 1.061e-02 & 1.757e-02 & 3.221e-02\\
        & SIREN & 5.180e-03 & 2.175e-01 & 2.436e-01 & 3.861e-01 & 8.995e-04 & 1.292e-02 & 1.783e-02 & 5.143e-02\\
        & \ours & \B 1.020e-04 & \B 4.504e-04 & \B 2.646e-04 & \B 5.951e-04 & \B 3.949e-06 & \B 4.436e-06 & \B 1.089e-05 & \B 1.174e-05 \\
        \midrule
        \multicolumn{10}{c}{$s=50\%$ subsampling} \\
        \midrule
        \multirow{4}{*}{\begin{turn}{90} In-s \end{turn}}
        & I-MP-PDE & \Uline{1.153e-04} & \Uline{5.016e-04} & \B 1.594e-04 & \Uline{6.043e-04} & \B 2.200e-07 & \Uline{3.179e-05} & \B 8.843e-07 & \Uline{5.854e-05} \\
        & DeepONet & 6.214e-04 & 4.277e-03 & 5.699e-03 & 1.082e-02 & 7.581e-04 & 1.187e-02 & 1.649e-02 & 3.378e-02 \\
        & SIREN & 4.911e-3 & 6.815e-1 & 1.607e-1 & 6.889e-1 & 5.134e-4 & 1.481e-2 & 3.086e-2 & 8.196e-2 \\
        & \ours & \B 8.151e-05 & \B 2.920e-04 & \Uline{2.004e-04} & \B 4.283e-04 & \Uline{3.277e-06} & \B 3.659e-06 & \Uline{8.978e-06} & \B 9.572e-06 \\
        \midrule
        \multirow{4}{*}{\begin{turn}{90} Out-s \end{turn}} & I-MP-PDE & \Uline{1.186e-04} & \Uline{5.010e-04} & \B 1.626e-04 & \Uline{6.132e-04} & \B 9.638e-07 & \Uline{3.153e-05} & \B 2.367e-06 & \Uline{5.574e-05} \\
        & DeepONet & 6.851e-04 & 4.343e-03 & 5.740e-03 & 1.099e-02 & 7.842e-04 & 1.185e-02 & 1.679e-02 & 3.391e-02 \\
        & SIREN & 5.067e-3 & 6.867e-1 & 1.599e-1 & 6.845e-1 & 5.354e-4 & 1.492e-2 & 3.113e-2 & 8.333e-2 \\
        & \ours & \B 9.175e-05 & \B 3.041e-04 & \Uline{2.116e-04} & \B 4.409e-04 & \Uline{3.277e-06} & \B 3.659e-06 & \Uline{8.978e-06} & \B 9.572e-06 \\
        \midrule
        \multirow{4}{*}{\begin{turn}{90} Full \end{turn}} & I-MP-PDE & \Uline{1.170e-04} & \Uline{5.013e-04} & \B 1.611e-04 & \Uline{6.088e-04} & \B 6.021e-07 & \Uline{3.166e-05} & \B 1.646e-06 & \Uline{5.710e-05} \\
        & DeepONet & 6.541e-04 & 4.311e-03 & 5.720e-03 & 1.091e-02 & 7.715e-04 & 1.186e-02 & 1.665e-02 & 3.385e-02 \\
        & SIREN & 4.995e-3 & 6.841e-1 & 1.603e-1 & 6.867e-1 & 5.246e-4 & 1.486e-2 & 3.100e-2 & 8.265e-2 \\
        & \ours & \B 8.677e-5 & \B 2.982e-4 & \Uline{2.062e-4} & \B 4.348e-4 & \Uline{3.380e-6} & \B 3.751e-6 & \Uline{9.251e-6} & \B 9.710e-6 \\
        \midrule
        \multicolumn{10}{c}{$s=100\%$ subsampling} \\
        \midrule
        \multirow{8}{*}{\begin{turn}{90} Full \end{turn}} & CNODE & 2.319e-2 & 9.652e-2 & 2.305e-2 & 1.143e-1 & 2.337e-05 & 5.280e-04 & 3.057e-05 & 7.288e-04 \\
        & MP-PDE & 1.140e-04 & \Uline{5.500e-4} & \B 1.785e-04 & \Uline{5.856e-04} & \B 1.718e-07 & \Uline{1.993e-05} & \B 9.256e-07 & \Uline{4.261e-5} \\
        & MNO & \B 3.190e-5 & 8.678e-4 & 2.763e-4 & 8.946e-4 & 9.381e-6 & 4.890e-3 & 1.993e-4 & 6.128e-3 \\
        & DeepONet & 1.375e-03 & 6.573e-03 & 9.704e-03 & 1.244e-02 & 6.431e-04 & 1.293e-02 & 1.847e-02 & 3.317e-02 \\
        & SIREN & 1.066e-3 & 4.336e-1 & 3.874e-1  & 1.037 & 3.674e-4 & 9.956e-3 & 3.013e-2 & 7.842e-2 \\
        & MFN & 1.651e-3 & 1.037e-0 & 2.106e-01 & 1.059e+00 & 1.408e-4 & 1.763e-1 & 4.735e-3 & 2.274e-1   \\
        & \ours (no sep.) & 3.235e-04 & 1.593e-03 & 7.850e-04 & 1.889e-03 & \Uline{2.641e-6} & 4.081e-5 & 5.977e-5 & 2.979e-4 \\
        & \ours & \Uline{8.339e-5} & \B 3.115e-4 & \Uline{2.092e-4} & \B 4.311e-4 & 3.309e-6 & \B 3.506e-6 & \Uline{9.495e-6} & \B 9.946e-6 \\
        \bottomrule
    \end{tabular}
\end{table}

\begin{table}
    \caption{\label{tab:results_new_grid_2_full} \textbf{Generalization across grids.} $\gX_{\tr}, \gX_{\ts}$ are subsampled with different ratios $s_{\tr}\neq s_{\ts}\in\{5, 50, 100\}$\% from the same uniform 64$\times$64 grid. We report \textit{test} MSE within $\gX_{ts}$ (\textit{In-s}). \textbf{Best} in bold.}
    \centering
    \sisetup{detect-weight, table-align-uncertainty=true, table-number-alignment=center, output-exponent-marker = \textsc{e}, table-format=1.3e-1, retain-zero-exponent,  exponent-product={}, mode=text}
    \scriptsize
    \begin{tabular}{llS@{\quad~~}SS@{\quad~~}SS@{\quad~~}SS@{\quad~~}S}
         \toprule
          & & \multicolumn{2}{c}{$\gX_{\ts}=\gX_{\tr}$} & \multicolumn{6}{c}{$\gX_{\ts}\neq\gX_{\tr}$} \\
        \cmidrule(lr){3-4} \cmidrule(lr){5-10}
        Subsampling & Test $\rightarrow$ & \multicolumn{2}{c}{$s_{\ts}=s_{\tr}$} & \multicolumn{2}{c}{$s_{\ts}=5\%$} & \multicolumn{2}{c}{$s_{\ts}=50\%$} & \multicolumn{2}{c}{$s_{\ts}=100\%$} \\
          \cmidrule(lr){3-4} \cmidrule(lr){5-6} \cmidrule(lr){7-8} \cmidrule(lr){9-10}
         Train $\downarrow$ & & {In-t} & {Out-t} & {In-t} & {Out-t} & {In-t} & {Out-t} & {In-t} & {Out-t} \\
         \midrule
          \multirow{2}{*}{$s_{\tr}=5\%$} & MP-PDE & \cellcolor{cgreen} \B 1.967e-04 & \cellcolor{cgreen} \B 6.631e-04  & \cellcolor{cred} 1.330e-01 & \cellcolor{cred} 3.852e-01 & \cellcolor{cred} 1.859e-01 & \cellcolor{cred} 6.680e-01 & \cellcolor{cred} 2.105e-01 & \cellcolor{cred} 7.120e-01 \\
          & \ours{} & \cellcolor{cgreen} 3.623e-04 & \cellcolor{cgreen} 8.306e-04 & \cellcolor{cyellow} \B 1.494e-03 & \cellcolor{cyellow}\B 2.291e-03 & \cellcolor{cyellow} \B 1.257e-03 & \cellcolor{cyellow} \B 1.883e-03 & \cellcolor{cyellow} \B 1.287e-03 & \cellcolor{cyellow} \B 1.947e-03 \\
         \midrule
         \multirow{2}{*}{$s_{\tr}=50\%$} & MP-PDE & \cellcolor{cgreen} \B 1.346e-04 & \cellcolor{cgreen} 5.110e-04 & \cellcolor{cred} 4.494e-02 & \cellcolor{cred} 9.403e-02 & \cellcolor{cyellow} 4.793e-03 & \cellcolor{cred} 1.997e-02 & \cellcolor{cyellow} 6.330e-03 & \cellcolor{cred} 3.712e-02  \\
         & \ours & \cellcolor{cgreen} 2.004e-04 & \cellcolor{cgreen} \B 4.283e-04 & \cellcolor{cgreen} \B 2.470e-04 & \B \cellcolor{cgreen} 4.697e-04 & \cellcolor{cgreen} \B 2.073e-04 & \cellcolor{cgreen} \B 4.284e-04 & \cellcolor{cgreen} \B 2.058e-04 & \cellcolor{cgreen} \B 4.361e-04 \\
        \midrule
        \multirow{2}{*}{$s_{\tr}=100\%$} & MP-PDE & \cellcolor{cgreen} \B \cellcolor{cgreen} 1.785e-04 & \cellcolor{cgreen} 5.856e-04 & \cellcolor{cred} 1.358e-01 & \cellcolor{cred} 3.355e-01 & \cellcolor{cred} 1.182e-02 & \cellcolor{cred} 2.664e-02 & \B \cellcolor{cgreen} 1.785e-04 & \cellcolor{cgreen} 5.856e-04 \\
         & \ours & \cellcolor{cgreen} 2.092e-4 & \cellcolor{cgreen} \B 4.311e-4 & \B \cellcolor{cgreen} 2.495e-04 & \B \cellcolor{cgreen} 4.805e-04 & \B \cellcolor{cgreen} 2.109e-04 & \cellcolor{cgreen} \B 4.325e-04 & \cellcolor{cgreen} 2.092e-4 & \cellcolor{cgreen} \B 4.311e-4 \\
         \bottomrule
    \end{tabular}
\end{table}

\section{Prediction}
\label{app:prediction}
\begin{figure}[H]
    \centering
    \resizebox{\textwidth}{!}{
    \setlength{\tabcolsep}{2pt}
    \footnotesize
    \begin{tabular}{ccC}
    \toprule
        \multirow{2}{*}{\makecell{Subsampling \\ rate}} & \multirow{2}{*}{\makecell{Observation \\ grid $\gX$}} & Predicted trajectory \\
        &  & ~~~$t=0$~~~\rightarrowfill~~~$t=T$~~~\rightarrowfill~~~$t=T'$~~~\\[0.1em]
        \midrule
        $s=5\%$ & \raisebox{-.415\totalheight}{\includegraphics[width=0.107\textwidth]{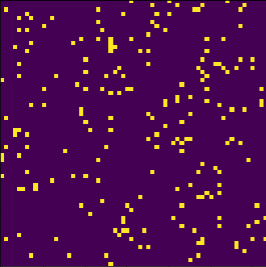}} & \includegraphics[width=0.8\textwidth]{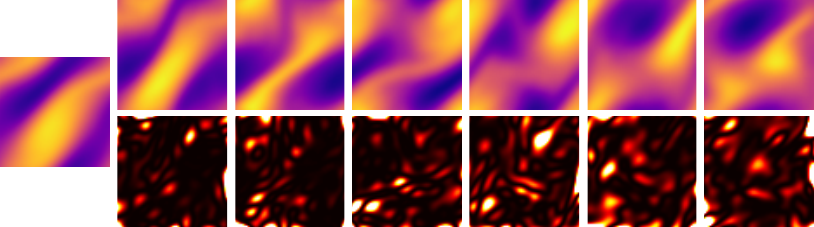} \\
         \midrule
        $s=25\%$ & \raisebox{-.415\totalheight}{\includegraphics[width=0.107\textwidth]{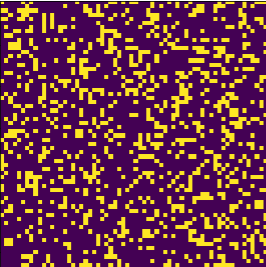}} & \includegraphics[width=0.8\textwidth]{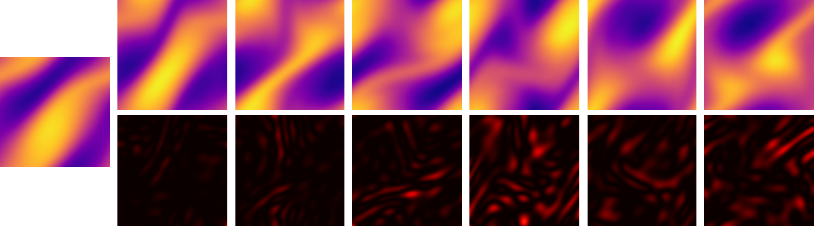}\\
        \midrule
        $s=100\%$ & \raisebox{-.415\totalheight}{\includegraphics[width=0.107\textwidth]{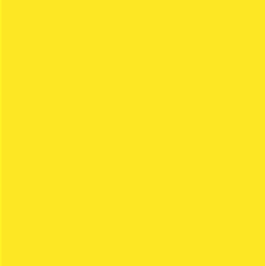}} & \includegraphics[width=0.8\textwidth]{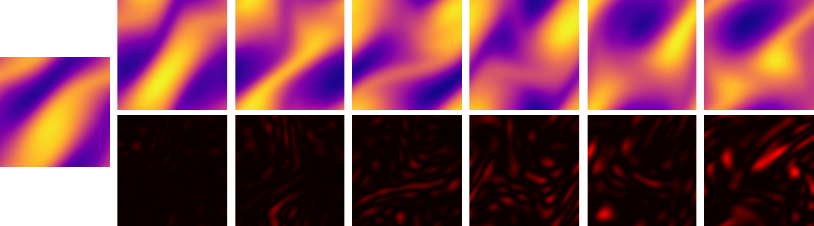} \\
        \midrule
        \multicolumn{2}{c}{Ground Truth} & \includegraphics[width=0.8\textwidth]{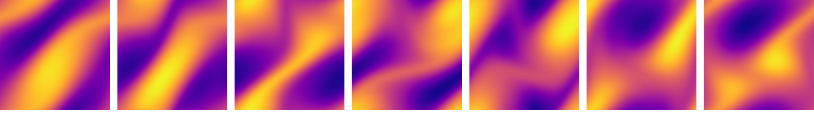} \\[-0.6em]
        \bottomrule
    \end{tabular}
    }
    \caption{\label{fig:ns_traj_dino} Prediction MSE per frame for \textbf{\ours} on \textit{Navier-Stokes} with its corresponding observed grid $\X$. For each model, the first row contains the predicted trajectory from $0$ to $T'$, the second row is the corresponding error maps \wrt the reference data (the darker the pixel, the lower the error).}
\end{figure}

\begin{figure}
    \centering
    \resizebox{\textwidth}{!}{
    \setlength{\tabcolsep}{2pt}
    \footnotesize
    \begin{tabular}{ccC}
    \toprule
        \multirow{2}{*}{\makecell{Subsampling \\ rate}} & \multirow{2}{*}{\makecell{Observation \\ grid $\gX$}} & Predicted trajectory \\
        &  & ~~~$t=0$~~~\rightarrowfill~~~$t=T$~~~\rightarrowfill~~~$t=T'$~~~\\[0.1em]
        \midrule
        $s=5\%$ & \raisebox{-.415\totalheight}{\includegraphics[width=0.107\textwidth]{figures/mask_5.png}} & \includegraphics[width=0.8\textwidth]{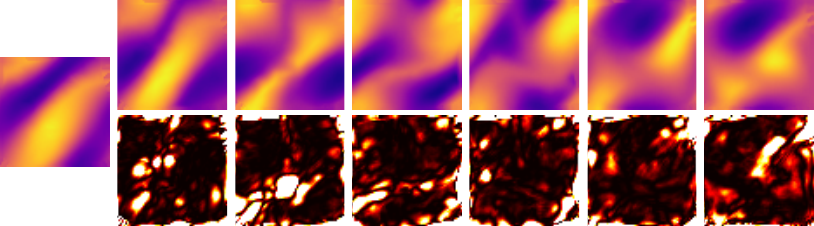} \\
         \midrule
        $s=25\%$ &  \raisebox{-.415\totalheight}{\includegraphics[width=0.107\textwidth]{figures/mask_25.png}} & \includegraphics[width=0.8\textwidth]{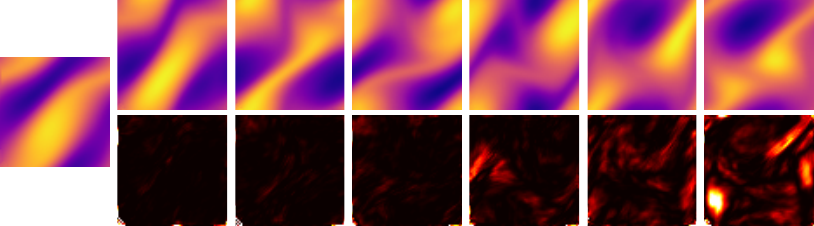} \\
        \midrule
        $s=100\%$ &  \raisebox{-.415\totalheight}{\includegraphics[width=0.107\textwidth]{figures/mask_100.png}} & \includegraphics[width=0.8\textwidth]{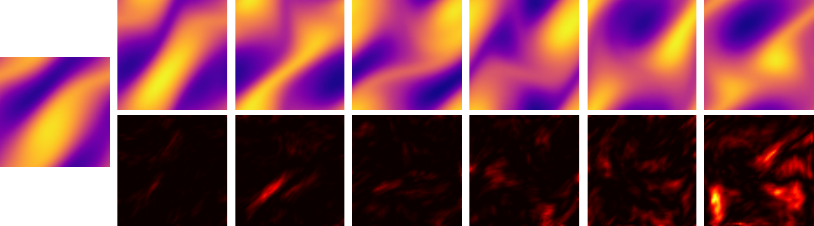} \\
        \midrule
        \multicolumn{2}{c}{Ground Truth} & \includegraphics[width=0.8\textwidth]{figures/visu_target_ns.pdf} \\[-0.6em]
        \bottomrule
    \end{tabular}
    }
    \caption{\label{fig:ns_traj_mppde} Prediction MSE per frame for \textbf{I-MP-PDE} on \textit{Navier-Stokes} with its corresponding observed grid $\X$. For each model, the first row contains the predicted trajectory from $0$ to $T'$, the second row is the corresponding error maps \wrt the reference data (the darker the pixel, the lower the error).}
\end{figure}

We display the test prediction of \ours (\Cref{fig:ns_traj_dino}) and I-MP-PDE (\Cref{fig:ns_traj_mppde}) for various subsampling levels when $\gX=\gX_{\tr}=\gX_{\ts}$.
Predictions are performed on a 64$\times$64 uniform grid which defines the observation grid $\gX$ via different subsampling rates.
Yellow points correspond to the observation grid $\gX$ (\textit{In-s}) while purple points indicate off-grid points (\textit{Out-s}).
The prediction for I-MP-PDE at $t=0$ is the interpolated initial condition.

\newcolumntype{C}{>{\centering\arraybackslash}m{11.5cm}}

\begin{figure}
    \centering
    \resizebox{\textwidth}{!}{
    \setlength{\tabcolsep}{2pt}
    \footnotesize
    \begin{tabular}{ccC}
    \toprule
        \multirow{2}{*}{\makecell{Subsampling \\ rate}} & \multirow{2}{*}{\makecell{Observation \\ grid $\gX$}} & Predicted trajectory \\
        &  & ~~~$t=0$~~~\rightarrowfill~~~$t=T$~~~\rightarrowfill~~~$t=T'$~~~\\[0.1em]
        \midrule
        $s=5\%$ & \raisebox{-.415\totalheight}{\includegraphics[width=0.107\textwidth]{figures/mask_5.png}} & \includegraphics[width=0.8\textwidth]{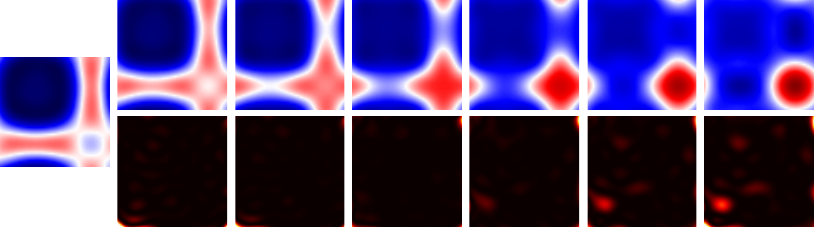} \\
         \midrule
        $s=25\%$ & \raisebox{-.415\totalheight}{\includegraphics[width=0.107\textwidth]{figures/mask_25.png}} & \includegraphics[width=0.8\textwidth]{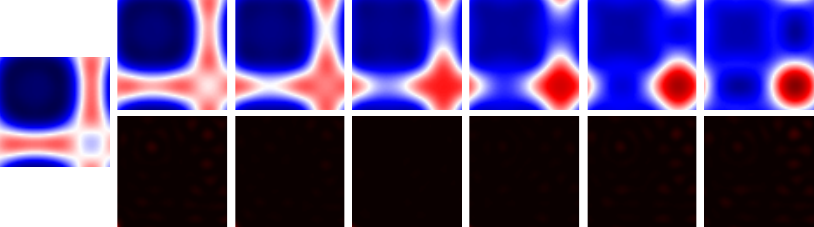}\\
        \midrule
        $s=100\%$ & \raisebox{-.415\totalheight}{\includegraphics[width=0.107\textwidth]{figures/mask_100.png}} & \includegraphics[width=0.8\textwidth]{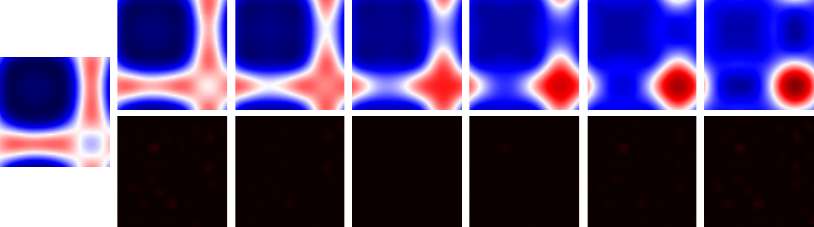} \\
        \midrule
        \multicolumn{2}{c}{Ground Truth} & \includegraphics[width=0.8\textwidth]{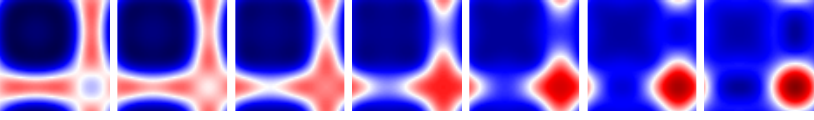} \\[-0.6em]
        \bottomrule
    \end{tabular}
    }
    \vspace{\baselineskip}
    \caption{\label{fig:we_traj_dino} \rev{Prediction MSE per frame for \textbf{\ours} on \textit{Wave} with its corresponding observed grid $\X$. For each model, the first row contains the predicted trajectory from $0$ to $T'$, the second row is the corresponding error maps \wrt the reference data (the darker the pixel, the lower the error).}}
\end{figure}

\section{Detailed description of datasets}
\label{app:data}
We choose $\gT$ (\resp $\gT'$) on a regular grid in $[0, T]$ (\resp $[0, T']$) with a given temporal resolution and fix $T'=2T + \delta t$, where $\delta t$ is the step size of the temporal grid.
Hence, we always consider 10 consecutive frames for \textit{In-t} and 10 more for \textit{Out-t}.
\rev{The range of $T$ depends on the nature of the dataset.}
We provide below further details on the choice of these parameters and other experimental parameters, such as the number of observed trajectories.

\paragraph{2D Wave equation \textnormal{(\textit{Wave})}.}
It is a second-order PDE:
\begin{equation}
    \dpd[2]{u}{t} = c^2\Delta u,
\end{equation}
where $u$ is a function of the displacement at each point in space \wrt the rest position, $c\in \R_+^*$ is the speed of wave traveling.
We transform the equation to a first-order form, considering the input $v_t = \parentheses*{u_t, \pd{u_t}{t}}$, so that the dimension of $v_t(x)$ at each point $x\in\Omega$ is $n=2$.

We generate our dataset for speed $c=2$ with periodic boundary condition.
The domain is $\Omega = [-1,1]^2$.
For initial conditions $v_0 = \Bigl(u_0, \rightvert*{\pd{u_t}{t}}_{t=0}\Bigr)$, the initial displacement $u_0$ is a Gaussian function:
\begin{equation}
    u_0(x;a,b,r)=a\app{\exp}{-{\frac{(x-b)^{2}}{2r^{2}}}},
\end{equation}
where the height of the peak displacement is $a\sim \gU(2,4)$, the location of the peak displacement is $(b_1, b_2)\sim \gU(-1,1)$, and the standard deviation is $r\sim \gU(0.25,0.3)$.
The initial time derivative is $\rightvert*{\frac{\partial u_t}{\partial t}}_{t=0} = 0$.
Each snapshot is generated on a uniform grid of 64$\times$64.
Each sequence is generated with fixed interval $\delta t=0.25$.
We set the train horizon $T=2.25$ and the inference horizon $T=4.75$.
We generated \num{512} train trajectories and \num{32} test trajectories.

\paragraph{2D Navier Stokes \textnormal{(\textit{Navier-Stokes}, \citealp{Stokes1851})}.} This dataset corresponds to an incompressible fluid dynamics described by:
\begin{equation}
    \frac{\partial w}{\partial t} = -u \nabla w + \nu \Delta w + f, \quad w = \nabla \times u, \quad \nabla u = 0,
\end{equation}
where $u$ is the velocity field and $w$ the vorticity.
$u, w$ lie on a spatial domain with periodic boundary conditions, $\nu$ is the viscosity and $f$ is a constant forcing term.
The input $v_t$ is $w_t$ ($n=1$). $\nu$ is the viscosity and $f$ is the constant forcing term in the domain $\Omega$.

The spatial domain is $\Omega = [-1,1]^2$, the viscosity is $\nu=\num{1e-3}$, the forcing term is set as:
\begin{equation}
    \forall x\in\Omega, f(x_1,x_2)= 0.1 \parentheses*{\app{\sin}{2 \pi (x_1 + x_2)} + \app{\cos}{2 \pi (x_1 + x_2)}}.
\end{equation}
The full spatial grid is of dimension 64$\times$64 or 256$\times$256 according to experiments in \Cref{sec:experiments}.
We sample initial conditions as in \citet{Li2021} to create different trajectories.
The first $20$ steps of the trajectories are cut off as they are too noisy and not informative in terms of dynamics.
Trajectories are collected with $\delta t = 1$. We set the training horizon $T=19$ and the inference horizon $T'=39$.
We generated \num{512} train trajectories and \num{32} test trajectories.

\paragraph{3D spherical shallow water \textnormal{(\textit{Shallow-Water}, \citealp{Galewsky2004})}.}
The following problem was originally presented for numerical model testing of global shallow-water equations. They can be written as:
\begin{equation}
    \begin{aligned}
        \dod{u}{t} &= -fk\times u- g\nabla h + \nu\Delta u, \\
        \dod{h}{t} &= -h\nabla\cdot u+\nu\Delta h.
    \end{aligned}
\end{equation}
where $\od{}{t}$ is the material derivative, $k$ is the unit vector orthogonal to the spherical surface, $u$ is the velocity field tangent to the surface of the sphere which can be transformed into the vorticity $w = \nabla\times u$, and $h$ is the thickness of the sphere. Note that the data we observe at each time $t$ is $v_t = (w_t, h_t)$. $f, g, \nu, \Omega$ are parameters of the Earth; \cf \citet{Galewsky2004} for details.

The initial conditions are slightly modified from \cite{Galewsky2004}, detailed below, to create symmetric phenomena on the northern and southern hemisphere. The initial zonal velocity $u_0$ contains two non-null symmetric bands in the both hemispheres, which are parallel to the circles of latitude. At each latitude and longitude $\phi, \theta \in [-\nicefrac{\pi}{2}, \nicefrac{\pi}{2}]\times[-\pi,\pi]$:
\begin{equation}
    u_0(\phi, \theta)=\begin{cases} \displaystyle
        \parentheses*{\frac{u_{\max}}{e_{\mathrm{n}}} \app{\exp}{\frac{1}{(\phi-\phi_0)(\phi-\phi_1)}}, 0} & \text{if~} \phi\in(\phi_0, \phi_1), \\
        \displaystyle \parentheses*{\frac{u_{\max}}{e_{\mathrm{n}}} \app{\exp}{\frac{1}{(\phi+\phi_0)(\phi+\phi_1)}}, 0} & \text{if~} \phi\in(-\phi_1, -\phi_0), \\
        (0, 0) & \text{otherwise.}
    \end{cases}
\end{equation}
where $u_{\max}$ is the maximum velocity, $\phi_0=\nicefrac{\pi}{7}, \phi_1=\nicefrac{\pi}{2}-\phi_0$, and $e_{\mathrm{n}} = \app{\exp}{-\nicefrac{4}{(\phi_1-\phi_0)^2}}$.
The water height $h_0$ is initialized by solving a boundary value condition problem as in \citet{Galewsky2004}. It is then perturbed by adding the following $h'_0$ to $h_0$:
\begin{equation}
    h'_0(\phi,\theta)=\hat{h} \cos (\phi) \app{\exp}{-\parentheses*{\frac{\theta}{\alpha}}^2} \brackets*{\app{\exp}{-\parentheses*{\frac{\phi_2-\phi}{\beta}}^2}+ \app{\exp}{-\parentheses*{\frac{\phi_2+\phi}{\beta}}^2}}.
\end{equation}
where $\phi_2=\nicefrac{\pi}{4}$, $\hat h = \SI{120}{\m}$, $\alpha = \nicefrac{1}{3}$ and $\beta = \nicefrac{1}{15}$ are constants defined in \cite{Galewsky2004}.

We simulate this phenomenon with Dedalus \citep{Burns2020} on a latitude-longitude (lat-lon) grid. The size of the grid is \num{128} (lat) $\times$ \num{256} (lon).
We take different initial conditions by sampling $u_{\max}\sim \gU(60,80)$ to generate long trajectories.
These long trajectories are then sliced into shorter ones.
For simulation, we take one snapshot per hour (of internal simulation time), \ie $\delta t =\SI{1}{\hour}$.
We stop the simulation at the $320$\up{th} hour.
To construct a dataset rich of dynamical phenomena, we take the snapshots within the last \SI{160}{\hour} in a long trajectory and slice them into $8$ shorter trajectoires.
Also note that the data is scaled into a reasonable range: the height $h$ is scaled by a factor of \num{3e3}, and the vorticity $w$ by a factor \num{2}.
In each short trajectory, $T=\SI{9}{\hour}$ and $T'=\SI{19}{\hour}$.
In total, we generated $16$ long trajectories (\ie $128$ short trajectories) for train, $2$ for test (\ie $16$ short trajectories).

\section{Implementation}
\label{app:implem}
We provide our code at \url{https://github.com/mkirchmeyer/DINo}.

\subsection{Algorithm}
\label{app:algo}
We detail the algorithm of \ours for training and test via pseudo-code in \Cref{alg:pdes}.
\rev{Training consists in solving \Eqref{eq:bi_level} \wrt $\psi, \alpha_\gT, \phi$. Inference involves optimization only to find $\alpha_0$.}

\begin{algorithm}
\SetKwInOut{Input}{Input}
\caption{\ours pseudo-code}\label{alg:pdes}
    \textit{\underline{Training}:} \Input{$\gD = \{v_{\gT}\}$, $\{\alpha^v_{\gT}\gets 0\}_{v\in\gD}$, $\phi \gets \phi_0$, $\psi \gets \psi_0$\;}
    \While{not converged}{
        \lFor*{$v\in\gD$}{
            $\alpha^{v}_{\gT} \gets \alpha^v_{\gT} - \eta_\alpha \nabla_{\alpha^v_{\gT}}\ell_{\text{dec}}(\phi, \alpha^v_{\gT})$\Comment*[r]{Modulation}
        }
        $\phi \gets \phi - \eta_\phi\nabla_{\phi}\left(\sum_{v\in\gD}\ell_{\text{dec}}(\phi, \alpha^v_{\gT})\right)$\Comment*[r]{Hypernetwork update}
        $\psi \gets \psi - \eta_\psi \nabla_{\psi}\left(\sum_{v\in\gD}\ell_{\text{dyn}}(\psi,\alpha^v_{\gT})\right)$\Comment*[r]{Dynamics update}
    }

    \textit{\underline{Test}:} \Input{$\gD' = \{v_{0}\}$, $\{\alpha^v_{0}\gets 0\}_{v\in\gD'}$, $\phi^\star$, $\psi^\star$, $\gT'\neq\gT$\;}
    \While{not converged}{
        \lFor*{$v\in\gD'$}{
            $\alpha^v_{0} \gets \alpha^v_{0} - \eta \nabla_{\alpha^v_{0}}\ell_{\text{dec}}(\phi^\star, \alpha^v_{0})$\Comment*[r]{Modulation}
        }
    }
    \For{$v\in\gD', t \in \gT'$}{
        $\alpha^v_{t} \gets \alpha^v_{0}+\int_{0}^{t} f_{\psi^\star}(\alpha^v_\tau)\dif\tau$\Comment*[r]{Unroll dynamics}
        $\tilde{v}_t \gets D_{\phi}(\alpha^v_{t})$\Comment*[r]{Predict}
    }
\end{algorithm}

\subsection{\rev{Convergence analysis}}
\label{app:convergence_analysis}

\begin{figure}
    \centering
    \includegraphics[width=\textwidth]{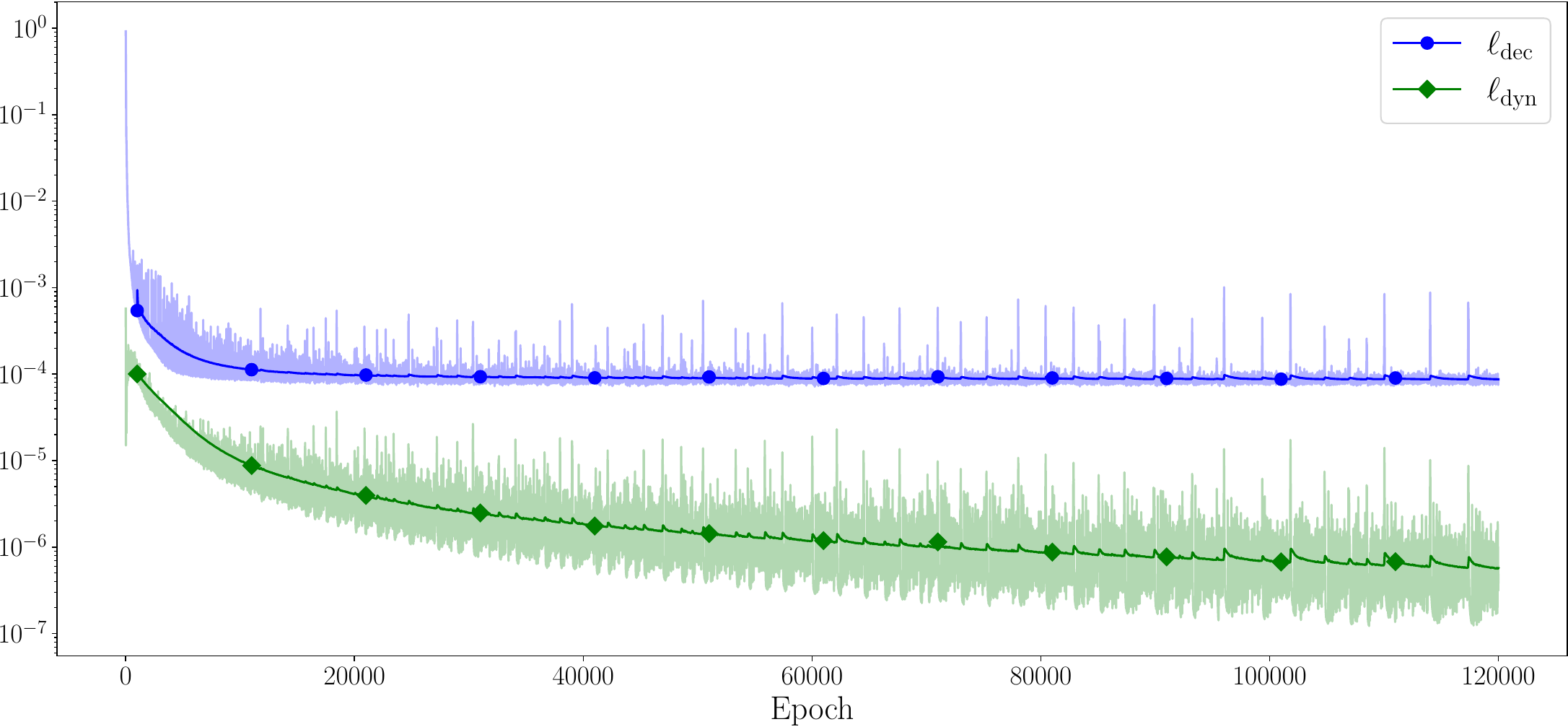}
    \caption{\label{fig:training_curves} \rev{Learning curves on \textit{Navier-Stokes} for $\ell_{\text{dec}}$ and $\ell_{\text{dyn}}$ throughout training (pale lines) and corresponding exponential moving averages from epoch \num{500} with half-life \num{1000} (opaque lines).}}
\end{figure}

\rev{In practice, we observe no training instability induced by the two-stage learning process of \Eqref{eq:bi_level} and \Cref{alg:pdes}: the objectives are non-conflicting. To assess this, we track the evolution of the auto-decoding loss $\ell_{\text{dec}}$ and the dynamics loss $\ell_{\text{dyn}}$ throughout training on \textit{Navier-Stokes} ($s=100$\%) in \Cref{fig:training_curves}. We observe that both losses smoothly converge until the end of training.}

\subsection{\rev{Time efficiency}}
\rev{Our auto-decoding strategy coupled with a latent neural ODE makes \ours computationally efficient compared to our best competitor MP-PDE. }

\paragraph{\rev{Inferring $\alpha_0$ via auto-decoding.}}
\rev{Given a decoder and an observation frame $v_0$, finding $\alpha_0$ corres\-ponds to solving an inverse problem, \cf \Eqref{eq:autodec}.
At inference, we use \num{300} steps to infer $\alpha_0$; using less steps is possible but results in slight underfitting.
This represents \SI{2.76}{\s} for $64$ trajectories on a single Tesla V100 Nvidia GPU.
Note that, as we unroll dynamics in the latent space, there is no need to relearn $\alpha_t$ when $t>0$.
Moreover, this differs from training, where $\alpha_t$ is continuously optimized for all $t\in[0,T]$ within the train horizon, in parallel with our INR decoder.
Overall, we trained MP-PDE and \ours for approximately $7$ days such that there is no major additional temporal training cost for \ours.}

\paragraph{\rev{Latent Neural ODE.}}
\rev{Unrolling the dynamics with a neural ODE is efficient (\SI{0.35}{\s} for $19$ time predictions for 64 trajectories on a single Tesla V100 Nvidia GPU).
Indeed, the latent space is small (at most $100$ dimension) and the dynamics model uses a simple four-layer MLP for $f_\psi$. With the same latent dynamics model, using an RK4 numerical scheme only incurs four additional function evaluations over a discretized alternative like a standard ResNet.
This incurs a minor computational cost but enables \ours to operate at different temporal resolutions, unlike \eg MP-PDE.}

\rev{In comparison, the official code of MP-PDE takes \SI{312}{\s} for inference on the same hardware for the same number of trajectories (vs \SI{3}{\s} for \ours). MP-PDE requires building an adjacency matrix and incurs for this reason a high memory cost, especially as the number of nodes increases. Interpolation also significantly increases inference time. This is not the case for \ours, which is faster.}

\subsection{Additional implementation details}
We use PyTorch \citep{Paszke2019} to implement \ours and our baselines.
\rev{Hyperparameters are further defined in \Cref{app:hyperparam}}.
The dynamics model $f_\psi$ is a multilayer perceptron with swish activation function \citep{Hendrycks2016, Ramachandran2017}.
Its input and output sizes are the same as the size of latent space $d_\alpha$.
All hidden layers share the same size.
\ours's parameters are initialized with the default initialization in PyTorch, \rev{defining $\phi_0, \psi_0, \omega$ in \Cref{alg:pdes}}.
\rev{We recall that $\omega$ is fixed throughout training to reduce the number of optimized parameters without loss of performance.}
\rev{As in related work \citep{Sitzmann2019,Fathony2021}}, the frequency parameters \rev{$\omega$} are scaled by a factor, \rev{$\omega_{\textrm{s}}$}, considered as a hyperparameter.
For dynamics learning, we use an RK4 integrator via \rev{TorchDiffEq \citep{Chen2018}} and apply exponential Scheduled Sampling \citep{Bengio2015} to stabilize training.
In practice, modulations $\alpha_t$ are learned channel-wise such that $I_\theta\colon\Omega\rightarrow \R^{d_c}$ has separate parameters per output dimension to make predictions less correlated across channels. We optimize all parameters \rev{$\phi, \alpha, \psi$} using Adam \citep{Kingma2015} with \rev{decay parameters} $(\beta_1,\beta_2)=(0.9, 0.999)$.

\subsection{Hyperparameters}
\label{app:hyperparam}
\begin{table}
    \centering
    \caption{\ours's hyperparameters.}
    \begin{tabular}{lccc}
    \toprule
       Hyperparameter & Navier-Stokes & Wave & Shallow-water \\
    \midrule
      \multicolumn{4}{c}{Decoder $D_\phi=I_{h_\phi}$}\\[0.2em]
      Number of layers & $3$ & $3$ & $6$ \\
      Number of hidden channels & $64$ & $64$ & $256$ \\
      Frequency scale factor \rev{$\omega_{\textrm{s}}$} & $64$ & $64$ & $64$ \\
      Size of latent space $d_\alpha$ & $100$ & $50$ & $300$ \\

    \midrule
      \multicolumn{4}{c}{Dynamics model $f_\psi$}\\[0.2em]
      Number of layers & $4$ & $4$ & $4$ \\
    Hidden layer size & $512$ & $512$ & $800$ \\
      Activation function & Swish & Swish & Swish \\
    \midrule
      \multicolumn{4}{c}{Optimization}\\
    Learning rate $\eta_\phi$ & $10^{-2}$ & $10^{-2}$ & $10^{-2}$ \\
      Learning rate $\eta_\alpha$ & $10^{-3}$ & $10^{-3}$ & $10^{-3}$ \\
    Learning rate $\eta_\psi$ & $10^{-3}$ & $10^{-3}$ & $10^{-3}$ \\
    Number\ of epochs & \num{12000} & \num{12000} & \num{12000} \\
    Batch size \ie sequences per batch & $64$ & $64$ & $16$ \\
    \bottomrule
    \end{tabular}
    \label{tab:hyperparam-dino}
\end{table}
We list the hyperparameters of \ours for each dataset in \Cref{tab:hyperparam-dino}.
In practice, we observe it is beneficial to decay the learning rates $\eta_\phi, \eta_\alpha$ when the loss reaches a plateau.

\rev{\subsection{Baselines implementation}
We detail in the following the hyperparameters and architectures used in our experiments for the considered baselines, which we reimplemented for our paper.
\begin{itemize}
    \item \textbf{CNODE} is implemented with four two-dimensional convolutional layers with $64$ hidden features, ReLU activations, $3\times3$ kernel and zero padding. Learning rate is fixed to $10^{-3}$. We use an adjoint method for integration like \citet{Chen2018}.
    \item \textbf{MNO.} We use the FNO architecture of \cite{Li2021} with three FNO blocks, GeLU activations, $12$ modes and a width of $32$. Learning rate is fixed to $10^{-3}$.
    \item \textbf{DeepONet.} We consider an autoregressive formulation of DeepONet. We choose a width of \num{1000} for hidden features with a depth of $4$ for both trunk and branch nets with ReLU activations. Learning rate is fixed to $10^{-5}$.
    \item \textbf{MP-PDE.} We adapt the implementation in \cite{Brandstetter2022} to handle two- and three-dimensional PDEs. We use a time window of $1$ with pushforward trick. Batch size and number of neighbors are fixed to $8$. Learning rate is fixed to $10^{-3}$. We use ReLU activations.
    \item \textbf{SIREN.} To represent data in space and time, SIREN takes space and time coordinates $(x,t)$ as input. To handle multiple trajectories, we concatenate an optimizable per-trajectory context code $\alpha$ to the coordinates like in \ours. We fix the hidden layer size of SIREN to $256$. We initialize the parameters and use the default input scale as in \cite{Sitzmann2019}. The size of the context code is $d_\alpha=800$. The learning rate is $10^{-3}$.
    \item \textbf{MFN.} Similarly to the previous SIREN baseline, we concatenate the per-trajectory context code to space and time coordinates at the first layer. The hidden layer size is fixed to $256$ and we use the default parameter initialization with a frequency scale $\omega_{\textrm{s}}$ of $64$ higher than \ours. The size of the context code is $d_\alpha=800$. The learning rate is $10^{-3}$.
\end{itemize}}

The ablation ``\ours (no sep.)'' modulates frequencies $\omega$s through a latent shift modulation from $\alpha_t$, similarly to the bias terms $b$s in \cref{sec:inr_archi}.

\section{\rev{Complementary analyses}}
\label{sec:complementary}

\rev{We detail in this section additional experiments, allowing us to further analyze and assess the performance of \ours.}

\subsection{\rev{Long-term temporal extrapolation}}
\label{app:long_term_extrapolation}
\rev{We provide in \cref{tab:long_term_extrapolation} an analysis of error accumulation over time for long-term extrapolation. More precisely, we generate a Navier-Stokes dataset with longer trajectories and report MSE for $T'=T+\Delta T$ where $\Delta T\in\{T, 5T, 10T, 50T\}$. Note that $\Delta T = T$ is the setting of our main experiments ($T' = 2T$).}

\rev{We observe that \ours's MSE in long-term forecasting is more than an order of magnitude smaller than for (I-)MP-PDE. This demonstrates the extrapolation abilities of our model. }

\begin{table}
    \caption{\rev{Long-term extrapolation performance of \ours and (I-)MP-PDE in the space and time generalization experiment for test trajectories on \textit{Out-t} ($] T, T'=T+\Delta T]$); \cf \cref{tab:results,sec:exp_setting}.}}
    \label{tab:long_term_extrapolation}
    \centering
    \sisetup{detect-weight, table-align-uncertainty=true, table-number-alignment=center, output-exponent-marker = \textsc{e}, table-format=1.3e-1, retain-zero-exponent, mode=text}
    \begin{tabular}{llSSSS}
        \toprule
        Subsampling ratio & Model & {$\Delta T=T$} & {$\Delta T=5 T$} & {$\Delta T=10 T$} & {$\Delta T=50 T$} \\
        \midrule
        \multirow{2}{*}{$s=5 \%$} & \ours & \B 2.017e-3 & \B 4.895e-3 & \B 1.209e-2 & \B 1.440e-1 \\
        & I-MP-PDE & 8.387e-3 & 3.580e-2 & 3.356e-1 & 4.031e+1 \\
        \multirow{2}{*}{$s=100 \%$} & \ours & \B 4.617e-4 & \B 2.082e-3 & \B 6.901e-3 & \B 1.215e-1 \\
        & MP-PDE & 5.251e-4 & 3.524e-2 & 3.339e-1 & 9.755e+1 \\
        \bottomrule
    \end{tabular}
\end{table}

\subsection{\rev{INRs' advantage over interpolation}}
\label{app:interpolation}
\rev{We report in \Cref{tab:interpol_inr} the MSE of bicubic interpolation, our FourierNet's MSE (auto-decoding with amplitude modulation but without dynamics model) and \ours's MSE (with dynamics model) on train \textit{In-t} for both \textit{Navier-Stokes} and \textit{Wave}. This corresponds to MSE averaged over all training frames within the train horizon and not only the initial condition $v_0$.}
\begin{table}
    \caption{\rev{MSE reconstruction error (In-s and Out-s) of train sequences within the train horizon (In-t) for three different methods: interpolation of observed points in $\gX_{\mathrm{tr}}$, FourierNet learned over individual frames in $\gX_{\mathrm{tr}}$, and \ours (FourierNet with a dynamics model).}}
    \label{tab:interpol_inr}
    \centering
    \sisetup{detect-weight, table-align-uncertainty=true, table-number-alignment=center, output-exponent-marker = \textsc{e}, table-format=1.3e-1, retain-zero-exponent, mode=text}
    \begin{tabular}{lSSS}
    \toprule MSE train In-t & {Interpolation} & {FourierNet} & {\ours} \\
    \midrule Navier-Stokes, $s=5 \%$ & 8.277e-3 & \B 9.673e-4 & 1.029e-3 \\
    Wave, $s=5 \%$ & 7.075e-4 & \B 4.085e-5 & 4.088e-5 \\
    \bottomrule
    \end{tabular}
\end{table}

\rev{We observe that FourierNet is better than interpolation. Indeed, interpolation is poorly adapted to sparse observation grids: the interpolation errors are clearly visible in \Cref{fig:ns_traj_mppde}, first row (5\% setting). Interestingly, \ours's MSE is only slightly worse than the FourierNet's MSE, showing that we correctly learned the dynamics of latent modulations $\alpha_t$. I-MP-PDE, which combines bicubic interpolation with MP-PDE, is then expectedly outperformed by \ours on this challenging 5\% setting. This shows the advantage of using INRs instead of standard bicubic interpolation to interpolate between observed spatial locations.}

\subsection{\rev{Modeling real-world data}}
\label{app:sst}

\paragraph{\rev{SST.}}
\rev{We evaluate \ours on real-world data to further assess its applicability. Following \cite{deBezenac2018} and \cite{Dona2021}, we model the Sea Surface Temperature (SST) of the Atlantic ocean, derived from the data-assimilation engine NEMO \citep[Nucleus for European Modeling of the Ocean,][]{Madec2008} using E.U.\ Copernicus Marine Service Information.\footnote{\url{https://data.marine.copernicus.eu/product/GLOBAL_ANALYSIS_FORECAST_PHY_001_024/description}.} Accurately modeling SST dynamics is critical in weather forecasting or planning of coastal activities. This problem is particularly challenging as SST dynamics are only partially observed: several unobserved variables affecting the dynamics (\eg the sea water flow) need to be estimated from data.}

\rev{For this experiment, we consider trajectories collected from three geographical zones (17 to 20) following the initial train\,/\,test split of \cite{deBezenac2018}. Notably, $T=\SI{9}{\day}$, which includes $\tau = \SI{4}{\day}$ of conditioning frames, \ie models are tested to predict $v_{t \in \lrbrackets*{\tau, T}}$ from $v_{t \in \lrbrackets*{0, \tau - 1}}$.}

\paragraph{\rev{Incorporating consecutive time steps.}}
\rev{To model SST which includes non-Markovian data and thus does not correspond to an Initial Value Problem as in \cref{sec:tasks}, we modify our dynamics model in a similar fashion to \cite{Yildiz2019} to integrate a history of several consecutive observations $v_{t \in \lrbrackets*{0, \tau - 1}}$ instead of only the initial observation $v_0$. In more details, we define a neural ODE over an augmented state $[\alpha_t, \alpha_t']$ where $\alpha_t$ is our auto-decoded state and $\alpha'_t$ is an encoding of $\tau=4$ past auto-decoded observations via a neural network $c_{\xi}$. We adjust our inference and training settings as follows:}
\begin{itemize}
    \item \rev{inference: we compute $\alpha'_{\tau-1}=c_{\xi}(\alpha_0, \ldots, \alpha_{\tau-1})$ and then unroll our neural ODE from the initial condition $[\alpha_{\tau-1}, \alpha'_{\tau-1}]$ to obtain $[\alpha_t, \alpha'_t]$ for all $t>\tau-1$:}
    \begin{align*}
        \rev{\forall t \in \lrbrackets*{0, \tau-1}, \alpha_{t}=e_{\varphi}(v_{t}),} && \rev{\alpha'_{\tau-1}=c_{\xi}(\alpha_0, \ldots, \alpha_{\tau-1}),} && \rev{\dod{[\alpha_t, \alpha'_t]}{t} = f_\psi ([\alpha_t, \alpha'_t]);}
    \end{align*}
    \item \rev{training: for all $t$, we infer $\alpha'_{t+\tau-1}=c_\xi(\alpha_t, \ldots, \alpha_{t+\tau-1})$ and fit the above neural ODE on the $[\alpha_t, \alpha'_t]$ obtained for all $t\in\lrbrackets*{0,T-\tau+1}$.}
\end{itemize}
\rev{This experiment confirms that our space- and time-continuous framework can easily be extended to incorporate refined temporal models. }

\begin{table}
    \caption{\rev{SST test prediction performance for \ours and VarSep \citep{Dona2021}.}}
    \label{tab:sst_mse}
    \centering
    \sisetup{detect-weight, table-align-uncertainty=true, table-number-alignment=center, mode=text}
    \begin{tabular}{lc}
        \toprule
        Method & MSE \\
        \midrule
        VarSep & 1.43 \\
        \ours & \B 1.27  \\
        \bottomrule
    \end{tabular}
\end{table}

\begin{figure}
    \centering
    \subfloat[\rev{Train predictions.}\label{tr_sst}]{\includegraphics[width=\linewidth]{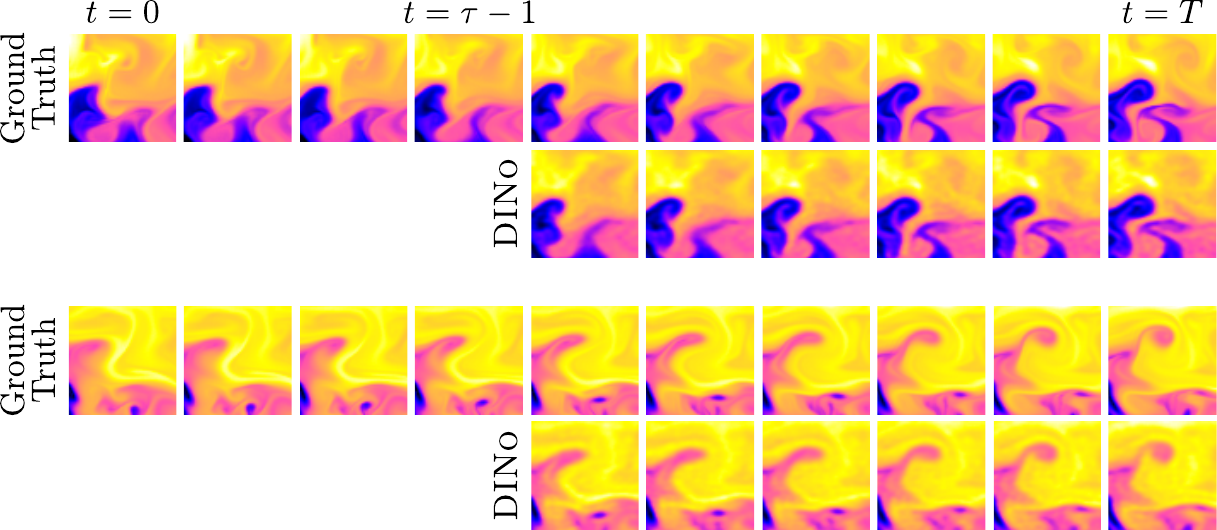}} \\
    \subfloat[\rev{Test predictions.}\label{ts_sst}]{\includegraphics[width=\linewidth]{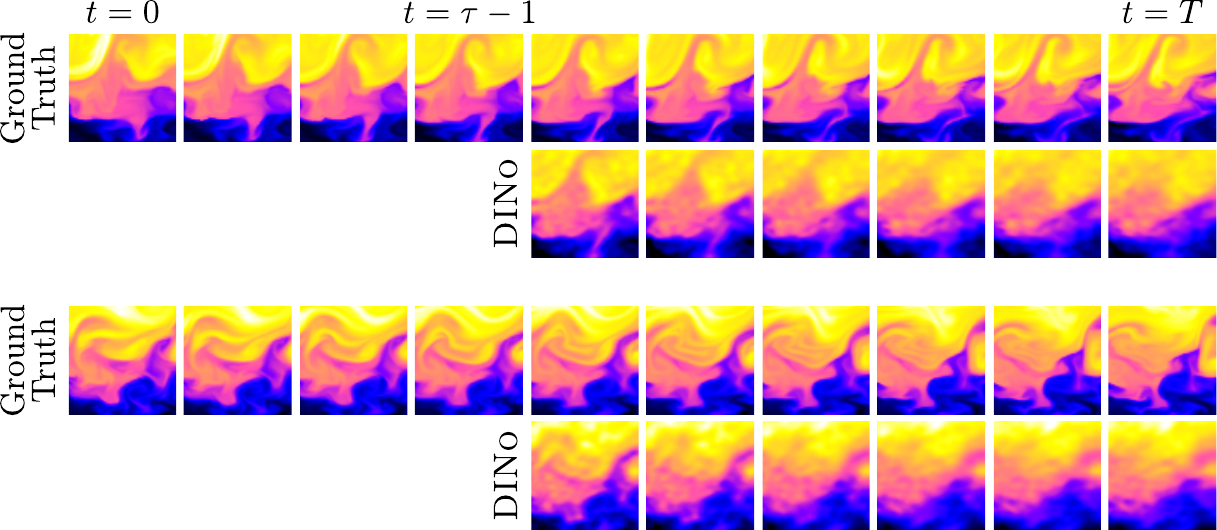}}
    \caption{\label{fig:results_sst}\rev{\ours's prediction examples on SST.}}
\end{figure}

\paragraph{\rev{Results.}}
\rev{ We report in \Cref{tab:sst_mse} test MSE for \ours and VarSep \citep{Dona2021}, the current state-of-the-art on SST, retrained on the same training data. \ours notably outperforms VarSep in prediction performance. This demonstrates \ours's potential to handle complex real-world spatiotemporal dynamics. We also provide some visualizations of \ours's train and test predictions in \Cref{fig:results_sst}. We make two observations. First, \ours fits very accurately the train data. Second, on the test data, we observe that the dynamics on low frequencies seem to be correctly modeled while the prediction of high frequencies dynamics are less accurate.
Larger scale experiments would be required to effectively evaluate the model performance on this challenging dataset. Given the complexity of the data, this is out of the scope of the paper.
Yet, these experiments already demonstrate that \ours behaves competitively \wrt the previous state-of-the-art. }

\paragraph{\rev{Implementation choices.}}
\rev{We choose a similar INR and dynamics architecture as for our Shallow-water experiment. We use for $c_\xi$, which takes as input four consecutive $\alpha_t$s, individual encoding of the $\alpha_t$s through a four-layer fully connected network which are then fed to a single linear layer.}

\end{document}